%% file: anonymous-submission-latex-2023.tex
\documentclass[letterpaper]{article} 
\usepackage{aaai23}
\usepackage{times}  
\usepackage{helvet}  
\usepackage{courier}  
\usepackage[hyphens]{url}  
\usepackage{graphicx} 
\usepackage{natbib}  
\usepackage{caption} 
\usepackage{xcolor}
\usepackage{algorithm}
\usepackage{algorithmicx}
\usepackage{algpseudocode}
\usepackage{newfloat}
\usepackage{listings}
\DeclareCaptionStyle{ruled}{labelfont=normalfont,labelsep=colon,strut=off} 
\floatstyle{ruled}
\newfloat{listing}{tb}{lst}{}
\floatname{listing}{Listing}
\usepackage[utf8]{inputenc} 
\usepackage{comment}
\usepackage{multirow}
\usepackage{booktabs}
\usepackage{longtable}
\usepackage{amsmath}
\usepackage{amssymb}
\usepackage{bbm}
\usepackage{caption}
\usepackage{subcaption}

\input{macro}

\definecolor{kscolor}{rgb}{0, 0, 0}
\definecolor{jycolor}{rgb}{0, 0, 0}
\newcommand{\ks}[1]{\textcolor{kscolor}{#1}}
\newcommand{\jy}[1]{\textcolor{jycolor}{#1}}

\definecolor{srcolor}{rgb}{0,0,0}

\newcommand*\samethanks[1][\value{footnote}]{\footnotemark[#1]}

\usepackage{hyperref}
\nocopyright

\title{
MIDMs: Matching Interleaved Diffusion Models\\ for Exemplar-based Image Translation
}
\author {
    Junyoung Seo \thanks{Equal contribution}\textsuperscript{\rm 1},
    Gyuseong Lee \samethanks \textsuperscript{\rm 1},
    Seokju Cho \textsuperscript{\rm 1},
    Jiyoung Lee \textsuperscript{\rm 2}
    Seungryong Kim \thanks{Corresponding author} \textsuperscript{\rm 1}
}
\affiliations {
    \textsuperscript{\rm 1} Korea University, Seoul, Korea \,
    \textsuperscript{\rm 2} NAVER AI LAB, Korea\\
}

\usepackage{bibentry}
\begin{document}

\maketitle

\begin{abstract}
 We present a novel method for exemplar-based image translation, called matching interleaved diffusion models (MIDMs). Most existing methods for this task were formulated as GAN-based matching-then-generation framework. However, in this framework, matching errors induced by the difficulty of semantic matching across cross-domain, e.g., sketch and photo, can be easily propagated to the generation step, which in turn leads to degenerated results. Motivated by the recent success of diffusion models overcoming the shortcomings of GANs, we incorporate the diffusion models to overcome these limitations. Specifically, we formulate a diffusion-based matching-and-generation framework that interleaves cross-domain matching and diffusion steps in the latent space by iteratively feeding the intermediate warp into the noising process and denoising it to generate a translated image. In addition, to improve the reliability of the diffusion process, we design a confidence-aware process using cycle-consistency to consider only confident regions during translation.
 Experimental results show that our MIDMs generate more plausible images than state-of-the-art methods. Project page is available at \url{https://ku-cvlab.github.io/MIDMs/}.
\end{abstract}

\input{1_introduction}

\input{2_related_works}

\input{3_preliminaries}

\input{4_methology}

\input{5_experiments}

\input{6_conclusion}

\bibliography{aaai23}
\input{appendix}

\end{document}

%% file: macro.tex
\newcommand{\eg}{\textit{e.g.}}
\newcommand{\ie}{\textit{i.e.}}

\def\eqref#1{Eq.(\ref{#1})}

\def\1{\bm{1}}

\DeclareMathAlphabet{\mathsfit}{\encodingdefault}{\sfdefault}{m}{sl}
\SetMathAlphabet{\mathsfit}{bold}{\encodingdefault}{\sfdefault}{bx}{n}

\def\gN{{\mathcal{N}}}

%% file: 1_introduction.tex
\begin{figure*}[t]
\centering
\begin{subfigure}[t]{0.378\textwidth}
    \includegraphics[width=\textwidth]{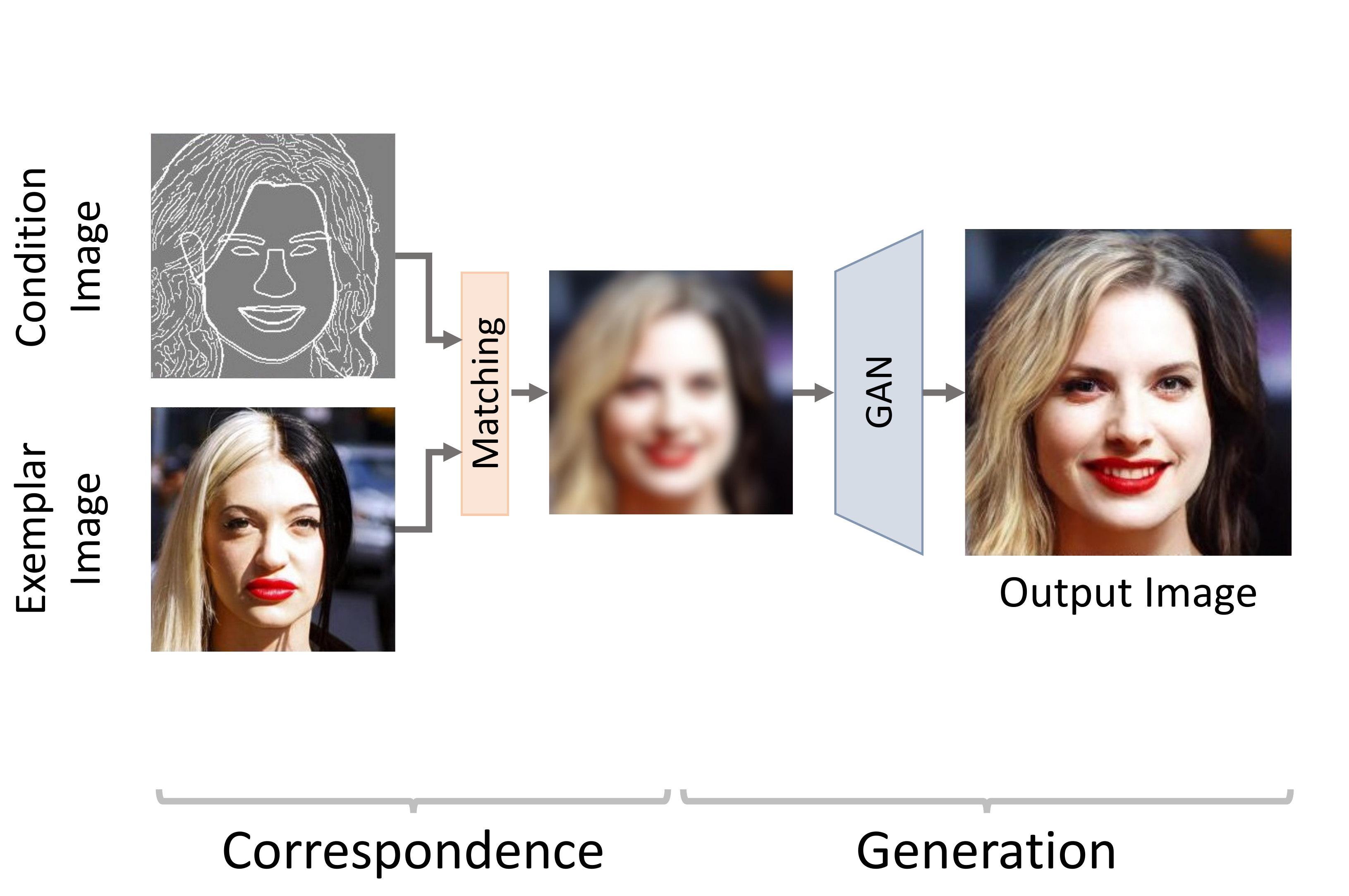}
    \caption{}
    \hfill
\end{subfigure}
\begin{subfigure}[t]{0.523\textwidth}
    \includegraphics[width=\textwidth]{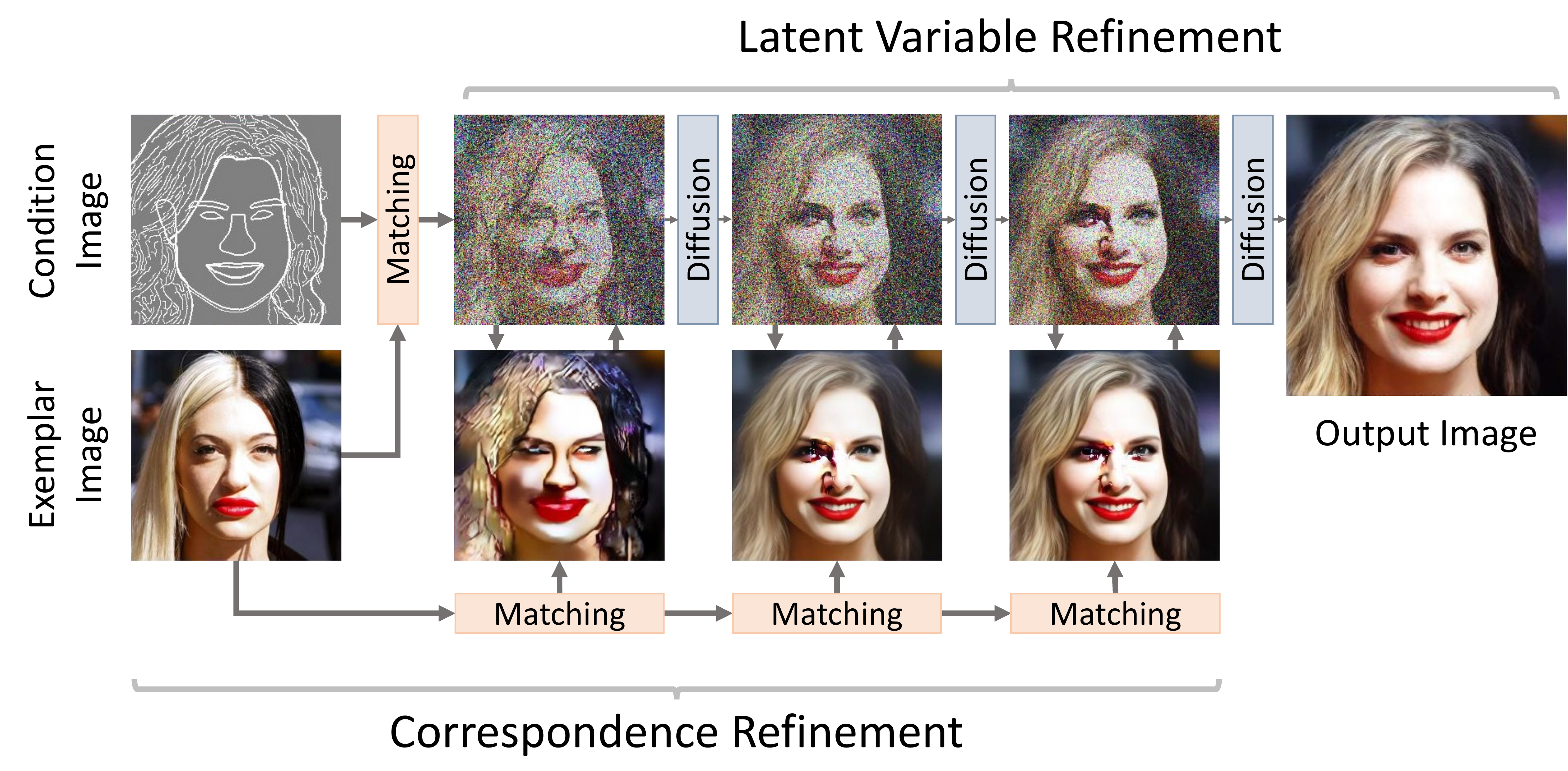}
    \caption{}
    \hfill
\end{subfigure}
\vspace{-10pt}\\
    \caption{\textbf{Motivation:} (a) existing works~\cite{Liao2017VisualAT, zhang2020cross, zhan2021unbalanced, zhan2021bilevelFA, zhou2021cocosnet, zhan2022marginal} and (b) our MIDMs include the interleaved process of the matching and generation, which can refine correspondence and embedded feature simultaneously.}
    \label{fig:attention}\vspace{-10pt}
\end{figure*}
\section{Introduction}

Image-to-image translation, aiming to learn a mapping between two different domains, has shown a lot of progress in recent years~\cite{zhu2017unpaired,isola2017image,Wang2018HighResolution,chen2017photographic,park2019semantic}. 
Especially, exemplar-based image translation~\cite{ma2018exemplar,wang2019example,zhang2020cross,zhou2021cocosnet,zhan2022marginal,zhan2021unbalanced} that can generate an image conditioned on an exemplar image has attracted much attention due to its flexibility and controllability.
For instance, translating a user-given condition image, e.g., pose keypoints, segmentation maps, or stroke, to a photorealistic image conditioned on an exemplar real image can be used in numerous applications such as semantic image editing or makeup transfer~\cite{zhang2020cross,zhan2021bilevelFA}.

To solve this task, early pioneering works~\cite{huang2018multimodal,ma2018exemplar,wang2019example} attempted to transfer a global style of exemplar. Recently, several works~\cite{zhang2020cross,zhou2021cocosnet,zhan2022marginal,zhan2021unbalanced} have succeeded in bringing the local style of exemplar by combining matching networks with Generative Adversarial Networks (GANs)~\cite{goodfellow2014generative}-based generation networks, i.e., GANs-based matching-then-generation. 
Formally, these approaches first establish matching across cross-domain and then synthesize an image based on a warped exemplar. However, the efficacy of such a framework is largely dependent on the quality of warped intermediates, which hinders faithful generations in case unreliable correspondences are established.
Furthermore, GANs-based generators inherit the weaknesses of the GAN model, i.e., convergence heavily depends on the choice of hyper-parameters~\cite{gulrajani2017improved,arjovsky2017wasserstein,salimans2016improvedtraininggans,goodfellow2016nips}, lower variety, and mode drop in the output distribution~\cite{brock2018large,miyato2018spectral}.
             

On the other hand, recently, diffusion models~\cite{sohl2015deep, ho2020denoising, song2020denoising, rombach2021high} have attained much attention as an alternative generative model. Compared to GANs, diffusion models can offer desirable qualities, including distribution coverage, a fixed training objective, and scalability~\cite{ho2020denoising,dhariwal2021diffusion,nichol2021glide}. Even though the diffusion models have shown appealing performances in image generation and manipulation tasks~\cite{choi2021ilvr,meng2021sdedit,kim2021diffusionclip}, applying this to exemplar-based image translation remains unexplored.


In this paper, we propose to use diffusion models for exemplar-based image translation tasks, called matching interleaved diffusion models (MIDMs), to address the limitations of existing methods~\cite{zhang2020cross,zhou2021cocosnet,zhan2021unbalanced,zhan2021bilevelFA,zhan2022marginal}. We for the first time adopt the diffusion models to exemplar-based image translation tasks, but directly adopting this in the matching-then-generation framework similarly to~\cite{zhang2020cross} may generate sub-optimal results. To overcome this, we present a diffusion-based matching-and-generation framework that interleaves cross-domain matching and diffusion steps to modify the diffusion trajectory toward a more faithful image translation, as shown in Fig.~\ref{fig:attention}. We allow the recurrent process to be confidence-aware by using the cycle-consistency so that our model can adopt only reliable regions for each iteration of warping.
The proposed MIDMs overcome the limitation of previous methods~\cite{zhang2020cross, zhou2021cocosnet, zhan2022marginal, zhan2021unbalanced} while transferring the detail of exemplars faithfully and preserving the structure of condition images.

Experiments demonstrate that our MIDMs achieve competitive performance on CelebA-HQ~\cite{liu2015deep} and DeepFashion~\cite{liu2016deepfashion}. 
In particular, user study and qualitative comparison results demonstrate that our method can provide a better realistic appearance while capturing the exemplar's details.
An extensive ablation study shows the effectiveness of each component in MIDMs.

%% file: 2_related_works.tex
\begin{figure*}[t]
\centering
\includegraphics[width=0.9\linewidth]{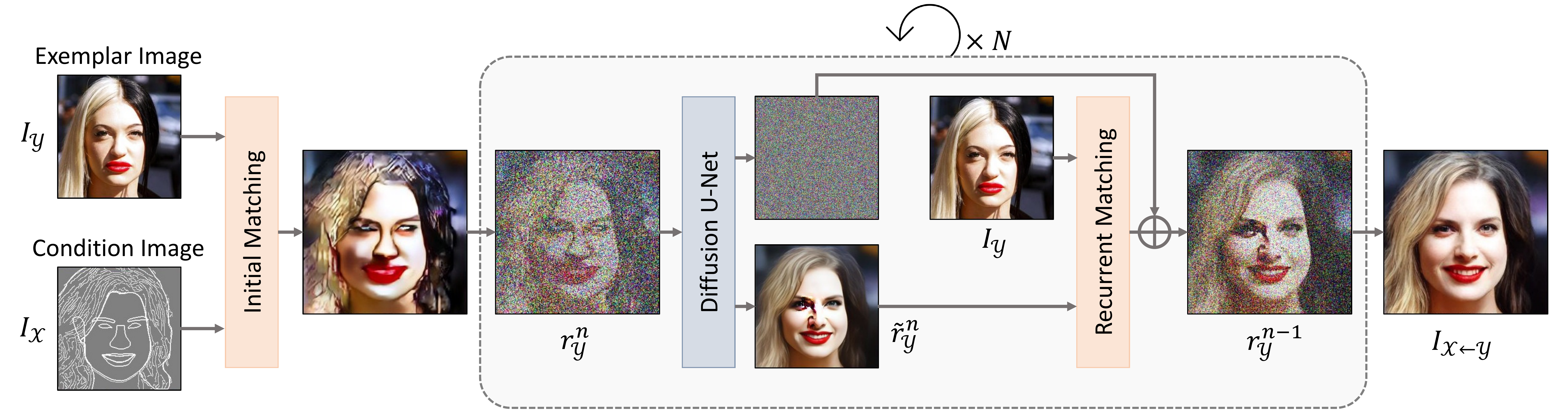}\hfill\\
\caption{\textbf{Overall architecture of MIDMs.} For condition image $I_\mathcal{X}$ and exemplar image $I_\mathcal{Y}$, we first compute initial matching and obtain the initial warped feature $\mathcal{R}_{\mathcal{X}\leftarrow{\mathcal{Y}}}$. Then we iteratively compute the diffusion and in-domain alignment with warped feature $r^{n}_{\mathcal{Y}}$ and reference $\mathcal{Y}$ to finally achieve $r^{0}_{\mathcal{Y}}$ that is used to achieve $I_{\mathcal{X}\leftarrow \mathcal{Y}}$.}
\label{fig:main_figure}\vspace{-10pt}
\end{figure*}
\section{Related Work}

\paragraph{Exemplar-based Image Translation.}
There have been a number of works~\cite{Bansal2019ShapesAC, wang2019example, Qi2018SemiParametricIS, huang2018multimodal} for exemplar-based image translation. 
Early works~\cite{huang2018multimodal} focused on bringing global styles, but recent works~\cite{Liao2017VisualAT, zhang2020cross, zhan2021unbalanced, zhan2021bilevelFA, zhou2021cocosnet, zhan2022marginal} have emerged to reference local styles by combining matching networks. 
While deep image analogy (DIA)~\cite{Liao2017VisualAT} proposed establishing dense correspondence, CoCosNet~\cite{zhang2020cross} suggested that building dense correspondence to cross-domain inputs makes the generated image preserve the given exemplar's fine details. Followed by this work, CoCosNet v2~\cite{zhou2021cocosnet} integrates PatchMatch~\cite{barnes2009patchmatch}. Although UNITE~\cite{zhan2021unbalanced} suggested unbalanced optimal transport~\cite{villani2009optimal} for feature matching to solve the many-to-one alignment problems, establishing feature alignment in cross-domain often fails because of domain gaps.
To solve this problem, MCL-Net~\cite{zhan2022marginal} introduced marginal contrastive loss~\cite{van2018representation} to explicitly learn the domain-invariant features.

\paragraph{Denoising Diffusion Probabilistic Models.}
Diffusion models generate a realistic image through the reverse of the noising process. 
With compelling generation results of many recent studies~\cite{ho2020denoising, dhariwal2021diffusion, nichol2021glide,rombach2021high, ramesh2022hierarchical}, diffusion models have emerged as a competitor to GAN-based generative models.
Recently, DDIM~\cite{song2020denoising} converted the sampling process to a non-Markovian process, enabling fast and deterministic sampling. Latent diffusion models (LDM)~\cite{rombach2021high} trained the diffusion model in a latent space by adopting a frozen pretrained encoder-decoder structure, which reduces computational complexity. 

Meanwhile, conditioning these diffusion models have been studied to make the controllable generation.
In SDEdit~\cite{meng2021sdedit}, proper amounts of noise were added to a drawing and denoised to recover the realistic image by the reverse process. DiffusionCLIP~\cite{kim2021diffusionclip} encodes the input image by the forward process of DDIM and finetunes the diffusion network with text-guided CLIP~\cite{Radford2021LearningTV} loss. However, there was no study to consider the connection between dense correspondence and image generation based on the diffusion models for exemplar-based image translation, which is the topic of this paper.

\paragraph{Correspondence Learning.}
Establishing visual correspondences enables building a dense correlation between visually or semantically similar images. Thanks to the rapid advance of convolutional neural networks (CNNs), many works~\cite{long2014convnets, Rocco2017ConvolutionalNN, Kim2017FCSSFC, Kim2018RecurrentTN, Cho2021SemanticCW, cho2022cats++} have shown promising results to estimate semantic correspondence.
Incorporating the correspondence model into the diffusion model is the topic of this paper.

%% file: 3_preliminaries.tex

\section{Preliminaries}
\paragraph{Diffusion Models.}
Diffusion models enable generating a realistic image from a normal distribution by reversing a gradual noising process~\cite{sohl2015deep, ho2020denoising}. Forward process, $q(\cdot)$, is a Markov chain that gradually converts to Gaussian distribution from the data $x_0 \sim q(x_0)$.
One step of forward process is defined as $q(x_t|x_{t-1}) :=\mathcal{N}(x_{t};\sqrt{1-\beta_t}x_{t-1}, \beta_t \mathbf{I})$, where $\beta_t$ is a pre-defined variance schedule in $T$ steps. The forward process can sample $x_t$ at an arbitrary timestamp $t$ in a closed form:
\begin{equation}
\begin{gathered}
    x_t = \sqrt{\alpha_t} x_0 + \sqrt{1 - \alpha_t} \epsilon, \\
    \alpha_t := \prod_{s=1}^{t} {(1-\beta_s)}, \quad \epsilon \sim \gN(0,\mathbf{I}).
    \label{eq:forward_ddpm}
\end{gathered}
\end{equation}
In addition, the reverse process is defined as $p_\theta(x_{t-1}|x_t):=\mathcal{N}(x_{t-1}; \mu_\theta (x_t, t), \sigma_\theta(x_t, t)\mathbf{I})$ that can be parameterized using deep neural network. DDPMs~\cite{ho2020denoising} found that using noise approximation model $\epsilon_\theta(x_t, t)$ worked best instead of using $\mu_\theta (x_t, t)$  to procedurally transform the prior noise into data.
Therefore, sampling of diffusion models is performed such that
\begin{equation}
    x_{t-1} = \frac{1}{\sqrt{1-\beta_t}}\left(x_t - \frac{\beta_t}{\sqrt{1-\alpha_t}} \epsilon_\theta(x_t, t)\right) + \sigma_t\epsilon.
    \label{eq:reverse_ddpm}
\end{equation}

\paragraph{Latent Diffusion Models.}
Recently, Latent Diffusion Models (LDM)~\cite{rombach2021high} reduces computation cost by learning diffusion model in a latent space. It adopts pretrained encoder $\mathcal{E}$ to embed an image to latent space and pretrained decoder $\mathcal{D}$ to reconstruct the image. In LDM, instead of $x$ itself, $z = \mathcal{E}(x)$ is used to define a diffusion process. Since DDIM~\cite{song2020denoising} uses an Euler discretization of some neural ODE~\cite{Chen2018NeuralOD}, enabling fast and deterministic sampling, LDM also adopted the DDIM sampling process. Intuitively, the DDIM sampler predicts $z_0$ directly from $z_t$ and then generates $z_{t-1}$ through a reverse conditional distribution. In specific, $f_\theta(z_{t}, t)$ is a prediction of $z_0$ given $z_t$ and $t$:
\begin{equation}\label{eq:f}
  f_\theta(z_{t}, t):= \frac{z_{t} - \sqrt{1-\alpha_t}\epsilon_{\theta}(z_t,t)}{\sqrt{\alpha_{t}}}.
 \end{equation}
The deterministic sampling process of DDIM in LDM is then as follows:
\begin{equation}
\begin{aligned}
    z_{t-1} = \sqrt{\alpha_{t-1}}f_\theta(z_{t}, t) +  \sqrt{1 - \alpha_{t-1} }{\epsilon}_{\theta}(z_{t}, t).
    \label{eq:ddim_original}
\end{aligned}
\end{equation}
After the diffusion process, an image is recovered such that $x = \mathcal{D}(z)$.

On the other hand, numerous works~\cite{saharia2021palette,rombach2021high} proposed a way to condition to the diffusion models. In specific, LDM proposes conditional generation by augmenting diffusion U-Net~\cite{Ronneberger2015UNetCN}. But these conditioning techniques cannot be directly applied to exemplar-based image translation tasks, which is the topic of this paper.

%% file: 4_methology.tex
\input{figure/fig_warping_process}
\section{Methodology}

\subsection{Problem Statement}
Let us denote a condition image and exemplar image as $I_\mathcal{X}$ and $I_\mathcal{Y}$, \eg, a segmentation map and a real image, respectively. 
Our objective is to generate an image $I_{\mathcal{X}\leftarrow \mathcal{Y}}$ that follows the content of $I_\mathcal{X}$ and the style of $I_\mathcal{Y}$, which is called an exemplar-based image translation task.

Conventional works~\cite{zhang2020cross,zhan2021unbalanced,zhan2022marginal} to solve this task typically followed two steps, cross-domain matching step between input images $I_\mathcal{X}$ and $I_\mathcal{Y}$, and image generation step from the warping hypothesis. Specifically, they first extract domain-invariant features $S_\mathcal{X}$ and $S_\mathcal{Y}$ from $I_\mathcal{X}$ and $I_\mathcal{Y}$, respectively, match them, and estimate an intermediate warp $R_{\mathcal{X}\leftarrow \mathcal{Y}}$ through the matches. An image generator, especially based on GANs~\cite{goodfellow2014generative}, then generates an output image $I_{\mathcal{X}\leftarrow \mathcal{Y}}$ from $R_{\mathcal{X}\leftarrow \mathcal{Y}}$. However, directly estimating cross-domain correspondence (e.g., sketch-photo) is much more complicated and erroneous than in-domain correspondence.
Thus they showed limited performance~\cite{zhan2021bilevelFA} depending on the quality of intermediate warp $R_{\mathcal{X}\leftarrow \mathcal{Y}}$.
In addition, they inherit the limitations of GANs, such as less diversity or mode drop in the output distribution~\cite{metz2016unrolled}.


\subsection{Matching Interleaved Diffusion Models (MIDMs)}

To alleviate the aforementioned limitations of existing works~\cite{zhang2020cross,zhan2021unbalanced,zhan2022marginal}, as illustrated in Fig.~\ref{fig:main_figure}, we introduce matching interleaved diffusion models (MIDMs) that interleave cross-domain matching and diffusion steps to modify the diffusion trajectory towards more faithful image translation, i.e., in a warping-and-generation framework. Our framework consists of cross-domain matching and diffusion model-based generation modules, that are formulated in an iterative manner. In the following, we first explain cross-domain matching and warping, diffusion model-based generation, and their integration in an iterative fashion. 

Similarly to LDM~\cite{rombach2021high}, we first define cross-domain correspondence and diffusion process in the intermediate latent space from pretrained frozen encoder-decoder, consisting of encoder $\mathcal{E}$ and decoder $\mathcal{D}$, so as to reduce the computation burden while preserving the image generation quality. 
In specific, given the condition image $I_\mathcal{X}$ and exemplar image $I_\mathcal{Y}$, we extract the embedding features $D_\mathcal{X}$ and $D_\mathcal{Y}$, respectively, through the pretrained encoder~\cite{Esser2021TamingTF} such that $D_\mathcal{X} = \mathcal{E}(I_\mathcal{X})$ and $D_\mathcal{Y} = \mathcal{E}(I_\mathcal{Y})$.
We abbreviate these as condition and exemplar respectively for the following explanations.


\paragraph{Cross-Domain Correspondence and Warping.}
For the cross-domain correspondence, our framework reduces a domain discrepancy by introducing two additional encoders, $\mathcal{F}_\mathcal{X}$ and $\mathcal{F}_\mathcal{Y}$ for condition and exemplar with separated parameters, respectively, to extract common features such that $S_\mathcal{X} = \mathcal{F}_\mathcal{X}(D_\mathcal{X})$ and $S_\mathcal{Y} = \mathcal{F}_\mathcal{Y}(D_\mathcal{Y})$. To estimate the warping hypothesis, we compute a correlation map $C_{\mathcal{X}\leftarrow \mathcal{Y}}$ defined such that
\begin{equation}
    C_{\mathcal{X}\leftarrow \mathcal{Y}}(u,v) = {S_\mathcal{X}(u)\over{\|S_\mathcal{X}(u)\|}} \cdot {S_\mathcal{Y}(v)\over{\|S_\mathcal{Y}(v)\|}},
    \label{eqn:cal-corr}
\end{equation}
where $u$ and $v$ index the condition and exemplar features, respectively. 

By taking the softmax operation, we can softly warp the exemplar $D_\mathcal{Y}$ according to $C_{\mathcal{X}\leftarrow \mathcal{Y}}$: 
\begin{equation}
    R_{\mathcal{X}\leftarrow \mathcal{Y}}(u)= \sum_v{\underset{v}{\mathrm{softmax}}(C_{\mathcal{X}\leftarrow \mathcal{Y}}(u,v)/\tau)} D_\mathcal{Y}(v),
    \label{eqn:warp}
\end{equation}
where $\tau$ is a temperature, controlling the sharpness of softmax operation.

\paragraph{Latent Variable Refinement Using Diffusion Prior.}
In this section, we utilize the diffusion process to refine the warped feature. Intuitively, given an initially-warped one, we add an appropriate amount of noise according to the standard forward process of DDPMs~\cite{ho2020denoising} to soften away the unwanted artifacts and distortions which may stem from unreliable correspondences, while preserving the structural information of the warped feature. Specifically, in the diffusion process, we feed $R_{\mathcal{X}\leftarrow \mathcal{Y}}$ to forward the process of DDPMs~\cite{ho2020denoising} to some extent and get the noisy latent variable $
r^{N}_{\mathcal{Y}}$ with proper $N$. We then iteratively denoise this, following an accelerated generation process in~\cite{song2020denoising}:
\begin{equation}
\begin{aligned}
  r^{n}_{\mathcal{Y}} = \begin{cases}
    \Tilde{r}^{1}_{\mathcal{Y}} \hfill (n=0) \\
  \sqrt{\alpha_{\tau_N}} R_{\mathcal{X}\leftarrow \mathcal{Y}}+\sqrt{1 - \alpha_{\tau_N}} \epsilon \hfill \qquad(n=N)\\
  \sqrt{\alpha_{\tau_n}} 
  \Tilde{r}^{n+1}_{\mathcal{Y}}+\sqrt{1 - \alpha_{\tau_n}} \epsilon_{\theta}(r^{n+1}_{\mathcal{Y}},\tau_{n+1}) \quad\hfill (o.w.)
  \end{cases} 
  \label{eqn:general-diffusion}
\end{aligned}
\end{equation}
 where $\Tilde{r}^{n}_{\mathcal{Y}} = f_{\theta}(r^{n}_{\mathcal{Y}},\tau_n)$, $\epsilon \sim \gN(0, \mathbf{I})$, and $o.w.$ means $otherwise$.
$\{\tau_n\}$ is a subsequence of time steps in the reverse process, i.e., the number of entire steps in the reverse process is reduced to $T$, which is the length of $\{\tau_n\}$. $N\in(0,T)$ is an intermediate step to initiate the reverse process.
By forwarding diffusion U-net~\cite{rombach2021high} and matching module iteratively, we get the refined latent variable $r^{0}_{\mathcal{Y}}$.



\paragraph{Interleaving Correspondence and Reverse Process.}
In this section, we explain how cross-domain correspondence is interleaved with denoising steps in an iterative manner. The intuition behind this is that matching the warped image and exemplar image is more robustly established than the matching between initial content and exemplar images as done in existing methods~\cite{zhang2020cross,zhan2021unbalanced,zhan2021bilevelFA,zhou2021cocosnet}. 
Specifically, we first feed the initially-warped exemplar $R_{\mathcal{X}\leftarrow \mathcal{Y}}$ to a noising process to get $
r^{N}_{\mathcal{Y}}$. We then feed it to \textit{one} step of sampling process to get a fully denoised prediction $\Tilde{r}^{N}_{\mathcal{Y}}$. Note that thanks to non-Markovian property of DDIM in Eq.~\ref{eq:f}, we can directly get a fully denoised prediction $\Tilde{r}^{N}_{\mathcal{Y}}$. In our framework interleaving correspondence and diffusion process, we intercept this, generate a better warped one, and then return to the denoising trajectory using the posterior distribution in Eq.~\ref{eq:ddim_original}.

In this framework, to achieve better correspondence at each step, 
we compute the correlation between $S_\mathcal{Y}$ and $\Tilde{r}^{n}_{\mathcal{Y}}$. To this end, we extract a feature defined such that 
\begin{equation}
    S^\mathrm{iter}_\mathcal{Y} = \mathcal{F}^\mathrm{iter}_\mathcal{Y}(\Tilde{r}^{n}_{\mathcal{Y}},D_\mathcal{X}),
    \label{eq:in_domain_feature}
\end{equation}
where $\mathcal{F}^\mathrm{iter}_\mathcal{Y}(\cdot)$ is a feature extractor designed for iteration, which receives refined warped exemplar $\Tilde{r}^{n}_{\mathcal{Y}}$ and condition $D_\mathcal{X}$ and mixes using a spatially-adaptive normalization~\cite{park2019semantic}. In fact, one can feed the  $\Tilde{r}^{n}_{\mathcal{Y}}$ to $\mathcal{F}_\mathcal{Y}$ instead of $\mathcal{F}^\mathrm{iter}_\mathcal{Y}$ since $\Tilde{r}^{n}_{\mathcal{Y}}$ is also from a real distribution. Nevertheless, as shown in~\cite{zhu2020sean}, we observe that injecting the condition $D_\mathcal{X}$ into the feature extractor can help to align the features and build more correct correspondences. We then compute a correlation map $C^\mathrm{iter}_{\mathcal{X}\leftarrow \mathcal{Y}}$ with $S_\mathcal{Y}$ and $S^\mathrm{iter}_\mathcal{Y}$ and extract the $R_{\Tilde{r}^{n}_{\mathcal{Y}}\leftarrow \mathcal{Y}}$. By returning $R_{\Tilde{r}^{n}_{\mathcal{Y}}\leftarrow \mathcal{Y}}$ to the denoising trajectory according to Eq.~\ref{eq:ddim_original}, we can obtain $r^{n-1}_{\mathcal{Y}}$. By iterating the above process, we finally obtain $r^{0}_{\mathcal{Y}}$. To summary, 
for $1 \leq n < N$, we change $\Tilde{r}^{n+1}_{\mathcal{Y}}$ in Eq.~\ref{eqn:general-diffusion} to $R_{\Tilde{r}^{n+1}_{\mathcal{Y}}\leftarrow \mathcal{Y}}$ as follows:
\begin{equation}
r^{n}_{\mathcal{Y}} = \sqrt{\alpha_{\tau_n}} 
  R_{\Tilde{r}^{n+1}_{\mathcal{Y}}\leftarrow \mathcal{Y}} + \sqrt{1 - \alpha_{\tau_n}} \epsilon_{\theta}(r^{n+1}_{\mathcal{Y}},\tau_{n+1}).
\label{eq:rewarp}
\end{equation}

\paragraph{Confidence-Aware Matching.}
There is a trade-off between bringing the details of exemplar faithfully and generating an image that matches the condition image, e.g., in the case of the condition image having earrings that do not exist in exemplar image~\cite{zhang2020cross}. To address this problem, we additionally propose a confidence-based masking technique. Specifically, we utilize a cycle-consistency~\cite{Jiang2021COTRCT} as the matching confidence at each warping step. We define the confidence mask such that
\begin{equation}
\begin{aligned}
  M_{\Tilde{r}^{n}_{\mathcal{Y}}\leftarrow \mathcal{Y}}(u) = \mathbbm{1}(\|u-\psi_{\mathcal{Y}\leftarrow \Tilde{r}^{n}_{\mathcal{Y}}}(\psi_{\Tilde{r}^{n}_{\mathcal{Y}}\leftarrow \mathcal{Y}} (u))\|_2^2<\gamma)
  \label{eq:mask}
\end{aligned}
\end{equation}
where $\psi$ is a warping function~\cite{Jiang2021COTRCT} and $\gamma$ is a threshold constant. Using this confidence mask $M_{\Tilde{r}^{n}_{\mathcal{Y}}\leftarrow \mathcal{Y}}$, we only rewarp the confident region and the rest region skips the rewarping process in Eq.~\ref{eq:rewarp} as
\begin{equation}
\begin{aligned}
    r^{n}_{\mathcal{Y}} = &\sqrt{\alpha_{\tau_n}} 
    (M_{\Tilde{r}^{n}_{\mathcal{Y}}\leftarrow \mathcal{Y}} \odot R_{\Tilde{r}^{n+1}_{\mathcal{Y}}\leftarrow \mathcal{Y}} + (1-M_{\Tilde{r}^{n}_{\mathcal{Y}}\leftarrow \mathcal{Y}}) \odot \Tilde{r}^{n+1}_{\mathcal{Y}}) \\+ &\sqrt{1 - \alpha_{\tau_n}} \epsilon_{\theta}(r^{n+1}_{\mathcal{Y}},\tau_{n+1}), 
\end{aligned}
\end{equation}
for $1 \leq n < N$. With this technique, the regions with low matching confidence intend to follow the reverse process of the general diffusion model. Intuitively, it allows selective control of the generative power depending on the matching confidence of the regions, which alleviates the aforementioned problem.

\paragraph{Image Reconstruction.}
Finally, we get the translated images by returning the latent variables to image space such that $I_{\mathcal{X}\leftarrow \mathcal{Y}} = \mathcal{D}(r^{0}_{\mathcal{Y}})$. We illustrate the whole process described above in Fig.~\ref{figure:editing}.

\subsection{Loss Functions}

Our model incorporates several losses to accomplish photorealistic image translation. 
\jy{Our core loss functions except for the diffusion loss  are similar to CoCosNet~\cite{zhang2020cross}.}
Note that we fine-tune the diffusion model with our loss functions.

\paragraph{Losses for Cross-Domain Correspondence.}
We use a pseudo-ground-truth image of a condition input image $I_\mathcal{X}$ as $I_\mathcal{X}'$.
We need to ensure that the extracted common features $\mathcal{S}_\mathcal{X}$ and $\mathcal{S}_\mathcal{X}'$ are in the same domain. 
\begin{equation}
\begin{aligned}
    \mathcal{L}_{\mathrm{dom}}=\|S'_\mathcal{X} - S_\mathcal{X}\|_1.
\end{aligned}
\end{equation}
In addition, the warped features should be cycle-consistent, which means that the exemplar needs to be returnable from the warped features. 
Because of our interleaved warping and generation process, we can acquire the cyclic-warped features at every $n$-th step:
\begin{equation}
    \mathcal{L}_{\mathrm{cycle}}=\sum\nolimits_{n}\|R_{\mathcal{Y}\leftarrow{\Tilde{r}^{n+1}_{\mathcal{Y}}}\leftarrow {\mathcal{Y}}} - D_\mathcal{Y}\|_1,
\end{equation}
where $R_{\mathcal{Y}\leftarrow{\Tilde{r}^{n+1}_{\mathcal{Y}}}\leftarrow {\mathcal{Y}}}$ is the cyclic-warped reference feature at n-step.

Finally, when we warp the ground-truth feature $\mathcal{D}_\mathcal{X}'$ with the correlation $\mathcal{C}_{\mathcal{X}\leftarrow{\mathcal{X}}'}$, we can obtain $R_{\mathcal{X}\leftarrow \mathcal{X}'}$, and this is consistent in terms of semantics, with the original ground-truth feature $\mathcal{D}_\mathcal{X}'$, building a source-condition loss $\mathcal{L}_{\mathrm{src}}$ as specified below:
\begin{equation}
    \mathcal{L}_{\mathrm{src}}=\| \phi_{l} (I_{\mathcal{X}\leftarrow{\mathcal{X}'}}) - \phi_{l} (I_\mathcal{X}')\|_{1}, 
\end{equation}
where $\phi_l$ is a $l$-th activation layer of pretrained VGG-19 model~\cite{Simonyan2015VeryDC}.

\paragraph{Losses for Image-to-image Translation.}
We use a perceptual loss~\cite{Johnson2016PerceptualLF} to maximize the semantic similarity since the semantic of the produced image should be consistent with the conditional input $I_\mathcal{X}$ or the ground truth $I_\mathcal{X}'$, denoted as follows:
\begin{equation}
\begin{aligned}
    \mathcal{L}_{\mathrm{perc}}=\|\phi_{l}(I_{\mathcal{X}\leftarrow{\mathcal{Y}}} )-\phi_{l} (I_\mathcal{X}')\|_1.
\end{aligned}
\end{equation}

Besides, we encourage the generated image $I_{\mathcal{X}\leftarrow{\mathcal{Y}}}$ to take the style consistency with the semantically corresponding patches from the exemplar $I_\mathcal{X}'$. Thus, we choose the contextual loss~\cite{mechrez2018contextual} as a style loss, expressed in the form of:
\begin{equation}
    \mathcal{L}_{\mathrm{style}}=-\log\left(\sum\nolimits_{l}\mu_{i}\mathrm{CX}_{ij}( \phi_{l} (I_{\mathcal{X}\leftarrow{\mathcal{Y}}}), \phi_{l} (I_\mathcal{Y})\right)
\end{equation}
where $\mathrm{CX}_{ij}$ is a contextual similarity function between images~\cite{mechrez2018contextual}.

\paragraph{Loss for Diffusion.}
We fine-tune a pretrained diffusion model~\cite{rombach2021high} that generates the high-quality outputs of the image domain. The diffusion objectives are defined as:
\begin{equation}
\begin{aligned}
    \mathcal{L}_{\mathrm{diff}}=\sum\nolimits_n{\| \epsilon_{\theta}(r^{n+1}_{\mathcal{Y}},\tau_{n+1}) - \epsilon\|_2},
\end{aligned}
\end{equation}
where $\epsilon$ is random noise used in the forward process of the diffusion~\cite{ho2020denoising}. \vspace{-5pt}

%% file: figure/fig_warping_process.tex

\begin{figure*}[t]
\centering
\begin{subfigure}[t]{0.150\textwidth}
    \includegraphics[width=\textwidth]{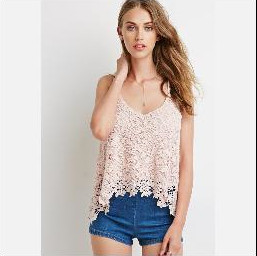}
    \caption{Exemplar}
\end{subfigure}
\begin{subfigure}[t]{0.150\textwidth}
    \includegraphics[width=\textwidth]{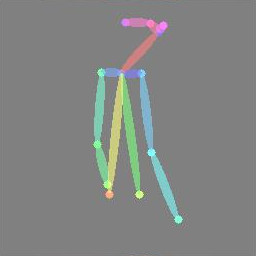}
    \caption{Condition}
\end{subfigure}
\begin{subfigure}[t]{0.150\textwidth}
    \includegraphics[width=\textwidth]{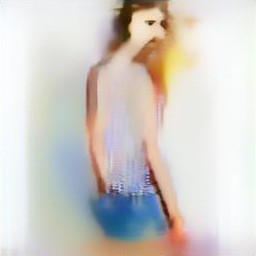}
    \caption{$n=5$}
\end{subfigure}
\begin{subfigure}[t]{0.150\textwidth}
    \includegraphics[width=\textwidth]{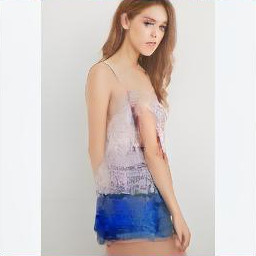}
    \caption{$n=4$}
\end{subfigure}
\begin{subfigure}[t]{0.150\textwidth}
    \includegraphics[width=\textwidth]{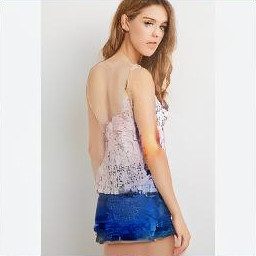}
    \caption{$n=2$}
\end{subfigure}
\begin{subfigure}[t]{0.150\textwidth}
    \includegraphics[width=\textwidth]{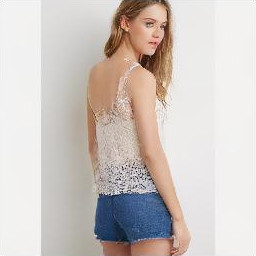}
    \caption{$n=0$}
\end{subfigure}
\vspace{-5pt}
\caption{\textbf{Examples of iterative matching-and-generation process:} (a) exemplar image, (b) condition image, (c)-(e) intermediate results of iterative process, which are refined gradually ($n=5, 4, 2$), and (f) final synthesis result ($n=0$).}
    \label{figure:editing} \vspace{-10pt}
\end{figure*}

%% file: 5_experiments.tex
\input{figure/fig_onecond}
\input{figure/fig_churches}

\input{table/main_table}
\input{table/main_style_table}

\section{Experiments}

\subsection{Experimental Settings}

\paragraph{Datasets.}
Following the previous literature~\cite{zhang2020cross,zhan2021bilevelFA,zhan2021unbalanced}, we conduct experiments over the CelebA-HQ~\cite{liu2015deep}, and DeepFashion~\cite{liu2016deepfashion} datasets.
CelebA-HQ~\cite{liu2015deep} dataset provides 30,000 images of high-resolution human faces at 1024×1024 resolution, and we construct the edge maps using Canny edge detector~\cite{canny1986computational} for conditional input.
DeepFashion~\cite{liu2016deepfashion} dataset consists of 52,712 full-length person images in fashion cloths with the keypoints annotations obtained by OpenPose~\cite{Cao2021OpenPoseRM}. 
\ks{Also, we use LSUN-Churches~\cite{yu2015lsun} to conduct the experiments of segmentation maps-to-photos. Because the LSUN-Churches dataset does not have ground-truth segmentation maps, we generate segmentation maps using Swin-S~\cite{liu2021swin} trained on ADE20k~\cite{zhou2017scene}. }

\paragraph{Implementation Details.}
We use AdamW optimizer~\cite{loshchilov2017decoupled} for the learning rate of 3e$-$6 for the correspondence network, and 1.5e$-$7 for the backbone network of the diffusion model. 
We use multi-step learning rate decay with $\gamma=0.3$.
We conduct our all experiments on RTX 3090 GPU, and we provide more implementation details and pseudo code in the Appendix. The codes and pretrained weights will be made publicly available.

\paragraph{Evaluation Metrics.}
To evaluate the translation results comprehensively, we report Fr{\'e}chet Inception Score (FID)~\cite{heusel2017gans} and Sliced Wasserstein distance~(SWD) to evaluate the image perceptual quality, \cite{karras2017progressive}, and Learned Perceptual Image Patch Similarity (LPIPS)~\cite{zhang2018unreasonable} scores to evaluate the diversity of translated images. 
Furthermore, we employ the style relevance and semantic consistency metrics~\cite{zhang2020cross} using a pretrained VGG model~\cite{Simonyan2015VeryDC}, which measures the cosine similarity between features of translated results and exemplar inputs.
Specifically, the low-level features (\ie, outputs of pretrained VGG network at $\texttt{relu}1\_2$ and $\texttt{relu}2\_2$ layers) are used to calculate color and style relevance, and high-level features (\ie, outputs of $\texttt{relu}3\_2, \texttt{relu}4\_2$ and $\texttt{relu}5\_2$ layers) are used to compute the semantic consistency score.

\subsection{Qualitative Evaluation}
Fig.~\ref{figure:qualitativeresultceleba} and Fig.~\ref{figure:qualitativeresultdeepfashion} demonstrate qualitative results with respect to different condition styles compared to CoCosNet~\cite{zhang2020cross}.
As can be seen therein, our method translates the detailed style of exemplar well in both datasets, preserving the structures of condition images.
{We also show diverse results on LSUN-Churces~\cite{yu2015lsun} in Fig.~\ref{figure:churches1}}. More qualitative results can be found in the Appendix.


\subsection{Quantitative Evaluation}
Table~\ref{tab_com} shows quantitative comparison with other exemplar-based image translation methods.
Thanks to the proposed interleaving cross-domain matching and diffusion steps, the proposed MIDMs outperform with large gaps in terms of SWD in both datasets.
Also in other metrics, our method demonstrates superior or competitive performance.
The semantic consistency and style consistency performance evaluations are summarized in Table~\ref{tab_consistency}. The proposed method achieves the best style relevance scores including both color and texture. \jy{We additionally evaluate the FID score compared with not only the distribution of source images as prior works~\cite{zhang2020cross,van2018representation} did, but also the distribution of exemplar images. In terms of FID compared with the distribution of exemplar images, MIDMs show superior results on all datasets we experiment with, which can be seen that our method translates the style of exemplar better. }


\input{figure/fig_user_study}
\subsection{User Study}
Finally, we conduct a user study to compare the subjective quality of the translated results in Fig.~\ref{fig:user_study}.
From 61 participants, we ask to rank all the methods in terms of style relevance and quality.
In all studies, we outperform the CoCosNet~\cite{zhang2020cross} and UNITE~\cite{zhan2021unbalanced}.




\subsection{Ablation Study}
We conduct ablation studies to demonstrate our model can find better correspondence generating more realistic images. Also, more ablation studies can be found in the Appendix.
\paragraph{Network Designs.}
From our best model, we validate our contribution by taking out the components of our model one by one in Table~\ref{table:ablation_network}. We observe a consistent decrease in performance when each component is removed. We can find that confidence masking is effective for our model. 
Replacing the recurrent matching process with one-time matching degrades the image quality significantly, which proves the superiority of our approach compared to the matching-then-generation framework.
\input{table/ablation_table_component}

\paragraph{Evaluations on Different Noise Levels.}
We also evaluate the FID score of our model for the different noise labels, and the results are shown in Table~\ref{table:ablation_noise}.
We observe that the proposed method with the 25\% noise level shows the best performance.
\input{table/ablation_table_noise}

\paragraph{Loss Functions.}
We conduct an ablation study to confirm the performance contribution of each loss function, by removing the loss term from our overall loss functions, and the result is shown in Table~\ref{table:ablation_loss}:
\input{table/ablation_table_loss}


%% file: figure/fig_onecond.tex
\begin{figure}[t]
\centering
\small
\setlength\tabcolsep{0pt}
{
\renewcommand{\arraystretch}{0.0}
\begin{tabular}{@{}rrrrrrr@{}}
    &
    \raisebox{0.32\height}{\rotatebox{90}{\scriptsize{Exemplars}}} \hspace{2pt}&
    \includegraphics[width=0.19\columnwidth]{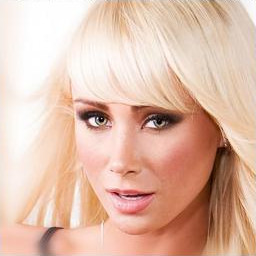}&
    \includegraphics[width=0.19\columnwidth]{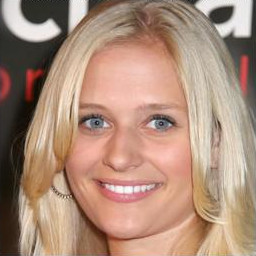}&
    \includegraphics[width=0.19\columnwidth]{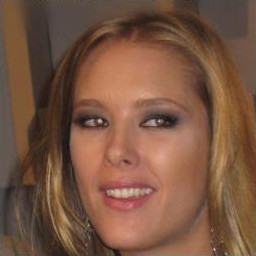}&
    \includegraphics[width=0.19\columnwidth]{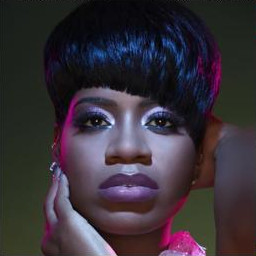}& \\
    \raisebox{0.27\height}{\rotatebox{90}{\scriptsize{CoCosNet}}}\hspace{2pt} &
    \includegraphics[width=0.19\columnwidth]{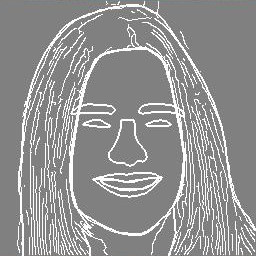}&
    \includegraphics[width=0.19\columnwidth]{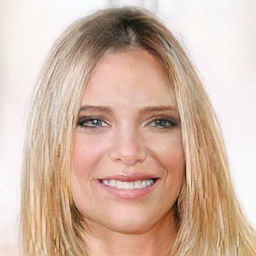}&
    \includegraphics[width=0.19\columnwidth]{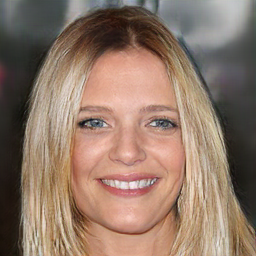}&
    \includegraphics[width=0.19\columnwidth]{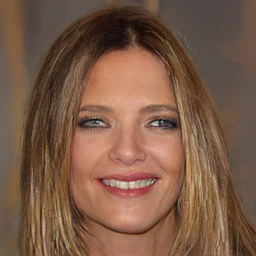}&
    \includegraphics[width=0.19\columnwidth]{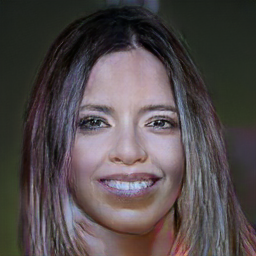}&     \\
    \raisebox{1.1\height}{\rotatebox{90}{\scriptsize{Ours}}} \hspace{2pt}&
    \includegraphics[width=0.19\columnwidth]{figure/Celeb/trg.jpg}&
    \includegraphics[width=0.19\columnwidth]{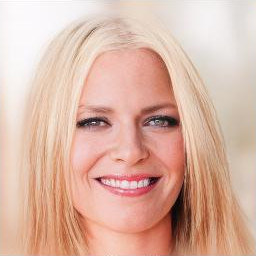}&
    \includegraphics[width=0.19\columnwidth]{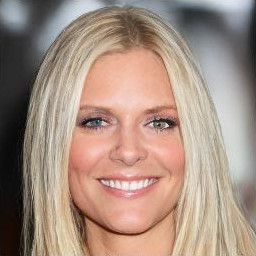}&
    \includegraphics[width=0.19\columnwidth]{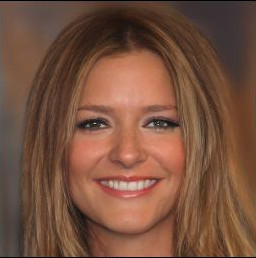}&
    \includegraphics[width=0.19\columnwidth]{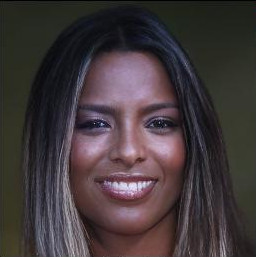}& \vspace{0.5em}    \\
\end{tabular}
}
\vspace{-5pt}
\caption{\textbf{Qualitative results for edge-to-face on CelebA-HQ~\cite{liu2015deep}:} (from top to bottom) exemplars, condition and results by CoCosNet~\cite{zhang2020cross} and our MIDMs.}
\label{figure:qualitativeresultceleba} \vspace{-10pt}
\end{figure}

\begin{figure}[t]
\centering
\small
\setlength\tabcolsep{0pt}
{
\renewcommand{\arraystretch}{0.0}
\begin{tabular}{@{}rrccccc@{}}
    &
    \raisebox{0.32\height}{\rotatebox{90}{\scriptsize{Exemplars}}} \hspace{2pt}&
    \includegraphics[width=0.19\columnwidth]{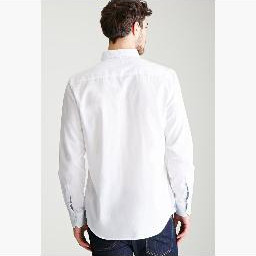}&
    \includegraphics[width=0.19\columnwidth]{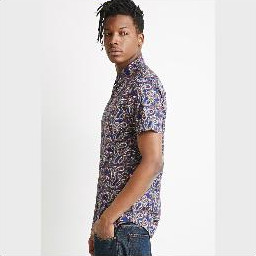}&
    \includegraphics[width=0.19\columnwidth]{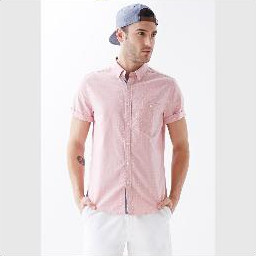}&
    \includegraphics[width=0.19\columnwidth]{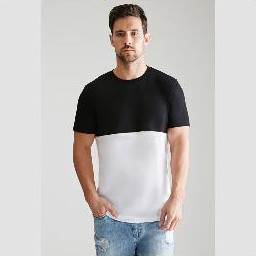}&
    \\
    \raisebox{0.27\height}{\rotatebox{90}{\scriptsize{CoCosNet}}}\hspace{2pt} &
    \includegraphics[width=0.19\columnwidth]{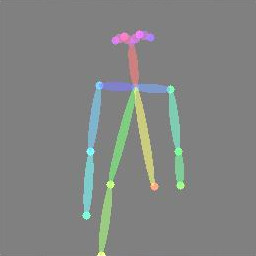}&
    \includegraphics[width=0.19\columnwidth]{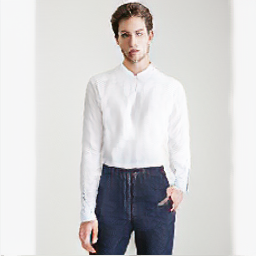}&
    \includegraphics[width=0.19\columnwidth]{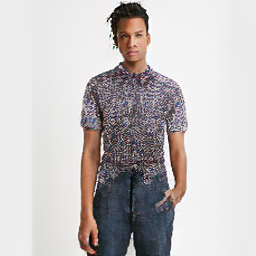}&
    \includegraphics[width=0.19\columnwidth]{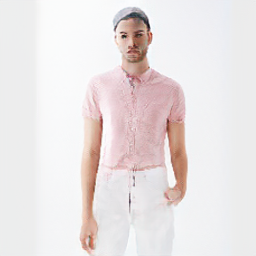}&
    \includegraphics[width=0.19\columnwidth]{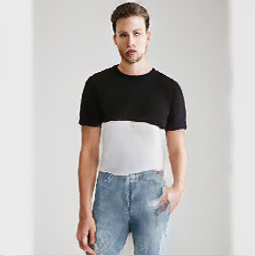}
    \\

    \raisebox{1.1\height}{\rotatebox{90}{\scriptsize{Ours}}} \hspace{2pt}&
    \includegraphics[width=0.19\columnwidth]{figure/Fashion/fashion_trg.jpg}&
    \includegraphics[width=0.19\columnwidth]{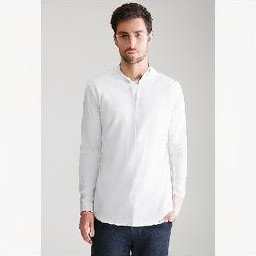}&
    \includegraphics[width=0.19\columnwidth]{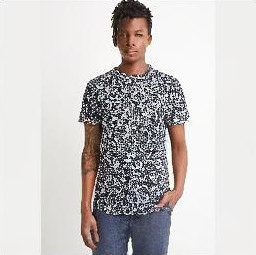}&
    \includegraphics[width=0.19\columnwidth]{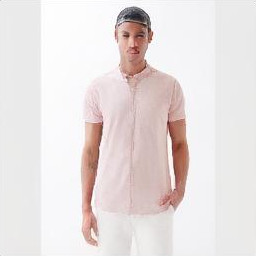}&
    \includegraphics[width=0.19\columnwidth]{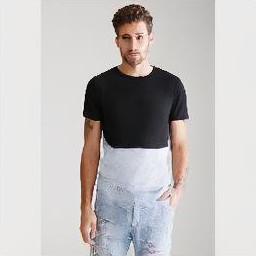}\vspace{0.5em}\\
\end{tabular}
}
\vspace{-5pt}
\caption{\textbf{Qualitative results for keypoints-to-photos on DeepFashion~\cite{liu2016deepfashion}:} (from top to bottom) exemplars, condition and results by CoCosNet~\cite{zhang2020cross} and our MIDMs.}
\label{figure:qualitativeresultdeepfashion} \vspace{-10pt}
\end{figure}

%% file: figure/fig_churches.tex
\begin{figure}[h]
\center
\small
\setlength\tabcolsep{0pt}
{
\renewcommand{\arraystretch}{0.0}
\begin{tabular}{@{}rrccccc@{}}
    &
    \raisebox{0.3\height}{\rotatebox{90}{\scriptsize{Exemplars}}} \hspace{15pt}&
    \includegraphics[width=0.19\columnwidth]{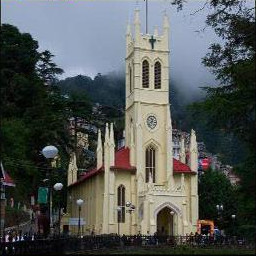}&
    \includegraphics[width=0.19\columnwidth]{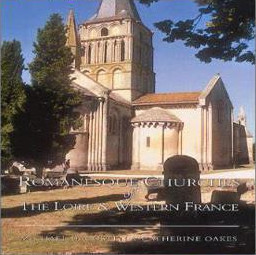}&
    \includegraphics[width=0.19\columnwidth]{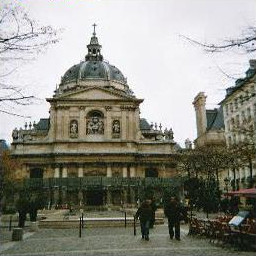}&
    \includegraphics[width=0.19\columnwidth]{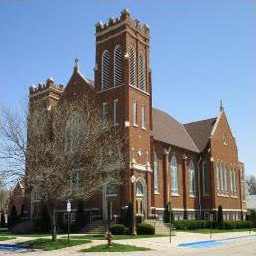}\\

    \raisebox{0.3\height}{\rotatebox{90}{\scriptsize{Condition}}} \hspace{15pt}&
    \includegraphics[width=0.19\columnwidth]{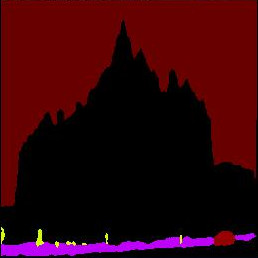}&
    \includegraphics[width=0.19\columnwidth]{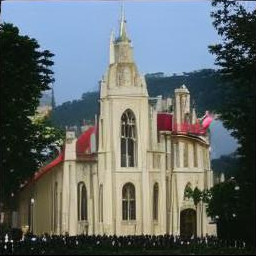}&
    \includegraphics[width=0.19\columnwidth]{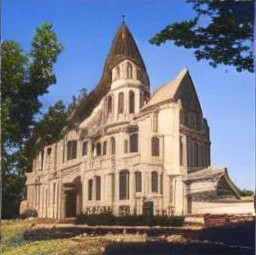}&
    \includegraphics[width=0.19\columnwidth]{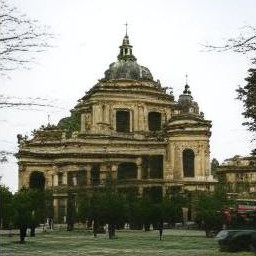}&
    \includegraphics[width=0.19\columnwidth]{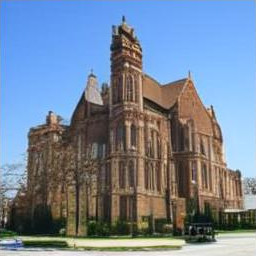}\\
\end{tabular}
}
\caption{\textbf{Qualitative results for segmentation maps-to-photos on LSUN-Churches~\cite{yu2015lsun}} }
\label{figure:churches1}
\end{figure}

%% file: table/main_table.tex
\renewcommand\arraystretch{1.0}

\begin{table*}[t]
\caption{
\textbf{Quantitative evaluation on DeepFashion and CelebA-HQ datasets.} The comparisons are performed with three widely used evaluation metrics FID~\cite{heusel2017gans}, SWD~\cite{karras2017progressive} and LPIPS~\cite{zhang2018unreasonable}. Some metrics for method that cannot be experimented because the codes or trained weights are not available are left blank. 
}\label{tab_com}
\vspace{-5pt}
\small
\centering 
\begin{tabular}{lcccccccc} 
\toprule
\multirow{3}[3]{*}{Methods}
& 
\multicolumn{4}{c}{DeepFashion~\cite{liu2016deepfashion} } &
\multicolumn{4}{c}{CelebA-HQ~\cite{liu2015deep}}
\\
\cmidrule(lr){2-5} \cmidrule(lr){6-9}
& \multicolumn{3}{c}{w/ sources } & w/ exemplars &
\multicolumn{3}{c}{w/ sources } & w/ exemplars
\\
& FID  $\downarrow$ & SWD  $\downarrow$ & LPIPS  $\uparrow$  & FID  $\downarrow$
& FID  $\downarrow$ & SWD  $\downarrow$ & LPIPS  $\uparrow$ & FID  $\downarrow$
\\
\midrule

Pix2pixHD~\cite{Wang2018HighResolution} & 25.20  & 16.40  & - & -       & 42.70 & 33.30 & - & -     \\

SPADE~\cite{park2019semantic} & 36.20  & 27.80  & 0.231 & -  & 31.50 & 26.90 & 0.187 & -       \\

SelectionGAN~\cite{tang2019multi}    & 38.31  & 28.21  & 0.223 & -     & 34.67 & 27.34 &  0.191 & -   \\

SMIS~\cite{zhu2020semantically} & 22.23  & 23.73  & 0.240 & -       &  23.71  &  22.23  &  0.201 & -   \\

SEAN~\cite{zhu2020sean}   & 16.28  &  17.52  & 0.251 & -       &  18.88    & 19.94 & 0.203 & -      \\

UNITE~\cite{zhan2021unbalanced}
& 13.08 & 16.65 & 0.278 & -
& 13.15 & 14.91 & 0.213 & -
  \\

CoCosNet~\cite{zhang2020cross}         & 14.40  & 17.20 & 0.272& 11.12        & 14.30 & 15.30 & 0.208 & 11.01      \\
CoCosNet v2~\cite{zhou2021cocosnet}        & 12.81  & 16.53 & 0.283& -        & 12.85 & 14.62 & 0.218 & -      \\ 

MCL-Net~\cite{zhan2022marginal}  & 12.89 & 16.24 & \textbf{0.286} & - & \textbf{12.52} & 14.21 & 0.216 & -  \\ 
\midrule
MIDMs (Ours)         & \textbf{10.89} & \textbf{10.10} & 0.279 & \textbf{8.54}  & 15.67 & \textbf{12.34} & \textbf{0.224} & \textbf{10.67} \\ 
\bottomrule
\end{tabular} 
\end{table*}

%% file: table/main_style_table.tex
\begin{table}[t]
\caption{\textbf{Quantitative evaluation of style relevance and semantic consistency on CelebA-HQ~\cite{liu2015deep}.}}
\label{tab_consistency}
\vspace{-5pt}
\small  
\centering 
\begin{tabular}{lccc} 
\toprule
\multirow{2}[3]{*}{Methods} & 
\multicolumn{2}{c}
{
Style relevance
} & 
\multirow{2}[3]{*}{\shortstack[c]{Semantic \\ consistency}}
\\
\cmidrule{2-3}
& Color & Texture &  \\
\midrule
Pix2PixHD~\cite{Wang2018HighResolution}    & -   &  -  & 0.914 \\

SPADE~\cite{park2019semantic}  &   0.955 &  0.927  & 0.922  \\


MUNIT~\cite{huang2018multimodal} & 0.939 & 0.884 &  0.848 \\

EGSC-IT~\cite{ma2018exemplar} & 0.965 & 0.942 &  0.915 \\

CoCosNet~\cite{zhang2020cross} & 0.977 & 0.958 &  \textbf{0.949}   \\
\midrule

MIDMs (Ours)  & \textbf{0.982} & \textbf{0.962} & 0.915    \\
\bottomrule

\end{tabular}
\end{table}

%% file: figure/fig_user_study.tex
\begin{figure}[t]
\centering
\includegraphics[width=0.4\textwidth]{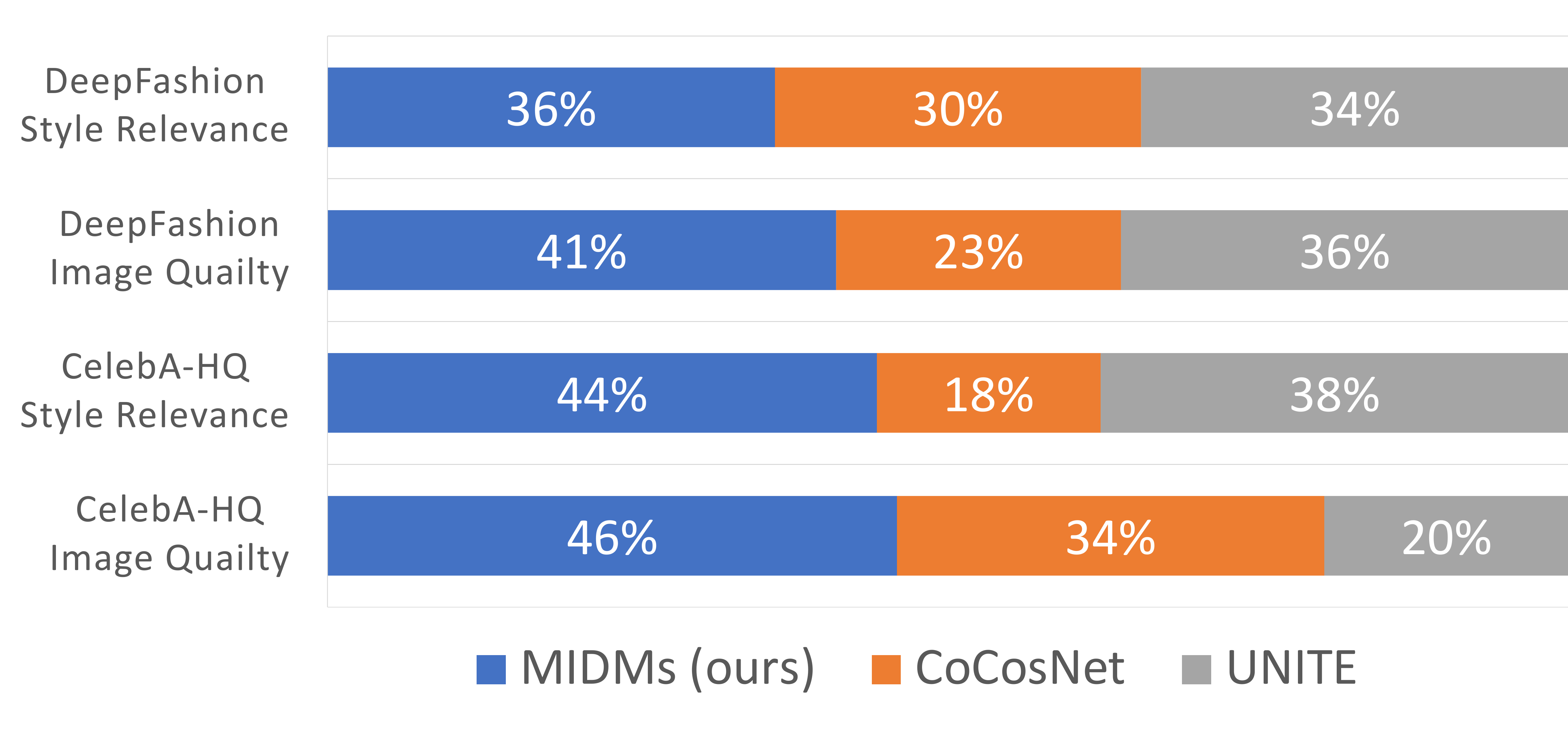}\vspace{-5pt}
\caption{\textbf{User studies on CelebA-HQ~\cite{liu2015deep} and DeepFashion~\cite{liu2016deepfashion}}.
\label{fig:user_study}} 
 \vspace{-10pt}
\end{figure} 

%% file: table/ablation_table_component.tex
\begin{table}[htb!]
\small
\centering
\caption{\textbf{Ablation study on the variants of components.} The baseline is our best model, and we validate the performance on CelebA-HQ~\cite{liu2015deep} by removing the elements one by one.}\label{table:ablation_network} \vspace{-5pt}
\begin{tabular}{lccc}
\toprule
{Models} & {FID}$\downarrow$ & {SWD}$\downarrow$ 
\\
\midrule
Ours & \textbf{15.67} &  \textbf{12.34}   \\
(-) Confidence Masking & 19.21 & 16.01    \\
(-) Recurrent Matching & 24.76 & 23.71   \\
(-) Diffusion U-Net        & 128.70 & 34.59  \\
\bottomrule
\end{tabular}



\end{table} 

%% file: table/ablation_table_noise.tex
\begin{table}[htb!]
\small
\centering
\caption{\textbf{Performance with respect to the noise levels at sampling.} 
We evaluate the performance on CelebA-HQ~\cite{liu2015deep}.} \label{table:ablation_noise}\vspace{-5pt} 
\begin{tabular}{lccc} 
\toprule
Noise & {FID}$\downarrow$ \\
\midrule
20\% & 23.67 &      \\
25\% & \textbf{15.67} \\
30\% & 16.01 \\
35\%  & 19.20  \\
\bottomrule
\end{tabular}
\end{table} 

%% file: table/ablation_table_loss.tex
\begin{table}[htb!]
\small
\centering
\caption{\textbf{Ablation study on each loss function.} We evaluate the performance on CelebA-HQ~\cite{liu2015deep}.} \label{table:ablation_loss}\vspace{-5pt} 
\begin{tabular}{lccc} 
\toprule
Loss & {FID}$\downarrow$ \\
\midrule
Ours & \textbf{15.67} &      \\
w/o $\mathcal{L}_{\mathrm{cycle}}$ & 16.18 \\
w/o $\mathcal{L}_{\mathrm{style}}$ & 19.23 \\
w/o $\mathcal{L}_{\mathrm{perc}}$  & 16.51  \\
w/o $\mathcal{L}_{\mathrm{dom}}$  & 16.68  \\
w/o $\mathcal{L}_{\mathrm{src}}$  & 72.25  \\
\bottomrule
\end{tabular}
\end{table} 

%% file: 6_conclusion.tex
\section{Conclusion}
In this paper, we presented MIDMs that interleave cross-domain matching and diffusion steps in the latent space by iteratively feeding the intermediate warp into the noising process and denoising it to generate a translated image.
To the best of our knowledge, it is the first attempt to use the diffusion models as a competitor to GANs-based methods in exemplar-based image translation.
Thanks to the joint synergy of the proposed modules, the style of exemplar were reliably translated to the condition input.
Experimental results show the superiority of our MIDMs for exemplar-based image translation as well as a general image translation task.

%% file: appendix.tex
\onecolumn
\section{Appendix}
In this document, we provide additional implementation details of MIDMs and more results.
\section{Additional Implementation Details}

\paragraph{Pretrained Encoder and Decoder.}
We use pretrained weights using a slightly modified encoder and decoder of VQGAN presented in~\cite{rombach2021high}. For CelebA-HQ~\cite{liu2015deep} and DeepFashion~\cite{liu2016deepfashion} dataset, we use VQ-regularized autoencoder with latent-space downsampling factor $f=4$. For LSUN-Churches~\cite{yu2015lsun} dataset we use KL-regularized autoencoder with latent-space downsampling factor $f=8$.
Therefore, the spatial dimension of latent space is $c \times (H/f)\times (W/f)$, where the channel dimension $c$ is $3$ for CelebA-HQ and DeepFashion and $4$ for LSUN-Churches dataset. Note that the encoder and the decoder are frozen and not finetuned.

\paragraph{Hyperparameters.}
We use almost identical hyperparameters of MIDMs on CelebA-HQ~\cite{liu2015deep} and DeepFashion~\cite{liu2016deepfashion} except masking threshold $\gamma$. $\lambda_{\mathrm{perc}}, \lambda_{\mathrm{src}}, \lambda_{\mathrm{style}}, \lambda_{\mathrm{cycle}}, \lambda_{\mathrm{dom}}, \lambda_{\mathrm{diff}}$ are scaling factors of $\mathcal{L}_{\mathrm{perc}}, \mathcal{L}_{\mathrm{src}}, \mathcal{L}_{\mathrm{style}}, \mathcal{L}_{\mathrm{cycle}}, \mathcal{L}_{\mathrm{dom}}, \mathcal{L}_{\mathrm{diff}}$, respectively. Please note that 4 out of 6 loss scaling factors are 1.0. Here, we provide a list of hyperparameters in Table~\ref{table:hyperparameter}.
\input{Table/table_hyperparameters}

\paragraph{Feature Refinement Techniques.} 
We find that skipping the warping process at the last step of warping-denoising iteration improves performance. Its implementation is trivial as we simply need to mask the entire area at the last step. Also, for generating realistic output, we refine the sampled feature again using the not-finetuned diffusion model by adding some noise back, which particularly improves the performance in terms of FID metric.

\paragraph{Warm-up Strategy.}
At the beginning of training, a warm-up strategy is used for good initialization of the correspondence network. Specifically, while the diffusion U-net is frozen, the correspondence network itself is trained only for the first 2 epochs. After the initial two epochs, all networks are trained in an end-to-end manner, except for the pretrained encoder and decoder, in a manner similar to ~\cite{rombach2021high}.

\paragraph{Sampling Details} 
We provide the pseudo code for MIDMs when sampling in Algorithm~\ref{alg:sample_alg}. Note that the iteration steps $N$ used for training can be greater in the sampling phase. The number of iteration steps $N$ used for sampling is 50.

\input{Table/sampling_algorithm}

\section{Additional Results}

\subsection{Analysis on Sampling Steps.}

As shown in~\cite{kim2021diffusionclip,ho2020denoising,song2020denoising}, diffusion models are known to show high image quality despite their speed slowing down as the number of sampling steps increases. Unlike the number of denoising steps during the training of MIDMs, the denoising steps in the sampling process can be further increased. In Table~\ref{table:sampling_steps}, we compare the performance when the sampling steps are 4, 25, and 50 respectively.

\input{Table/sampling_step_table}



\subsection{Applying the GANs to the proposed matching-and-generation framework.}
We apply the iterative matching-and-generation, our main idea, to the GAN-based models. In specific, we repeatedly feed the output of the generator, which is the generated image, to the matching network of the CoCosNet~\cite{zhang2020cross}. Because the conditional input is in the same domain as the reference image in this stage, we substitute the conditional input encoder with the photo-realistic input encoder and obtained the matching result. The last part is the same as the original CoCosNet~\cite{zhang2020cross}, which is the warped image-conditioned generation. Its qualitative results are shown in Table~\ref{figure:gan_abl}, and we also offer the quantitative result shown in Table~\ref{tab:gan_abl}.
We find that the GAN-based matching-and-generation produces worse results in both qualitative and quantitative, and as a result, we claim that applying GANs to this technique without training is not useful. The intuitions that we adopt the diffusion model and reasons for the result of the GAN-based ablation experiment we speculate are as follows:

First of all, one of the characteristics of the diffusion model we want to leverage is that an intermediate image in the middle of the generation can be explicitly extracted. GANs may also be applied to our proposed framework, but the diffusion model differs in that the matching process can be involved in the intermediate process of generation (i.e., in the middle of the trajectory approaching the real distribution from the prior distribution) rather than between complete generation results (i.e., already in the real distribution because GANs jump to real distribution at once thanks to implicit training with the discriminator). In fact, the method of imposing external guidance when the diffusion model gradually converts the image from prior distribution to target distribution has been verified to be effective in various works~\cite{dhariwal2021diffusion,ho2021classifier}, and this support why the diffusion model is adopted. Secondly, GAN-based exemplar-based I2I models inherit the weaknesses of GANs, i.e. the lack of mode coverage. Besides, diffusion models predict the likelihood distribution explicitly and tend to have a relatively large coverage of the distribution~\cite{dhariwal2021diffusion}.
\input{Figure/gan_abl}
\input{Table/gan_ablation}

\subsection{More Qualitative Results} 
We provide more generation results of MIDMs on CelebA-HQ~\cite{liu2015deep}, DeepFashion~\cite{liu2016deepfashion}, and LSUN-Churches~\cite{yu2015lsun} in Fig.~\ref{figure:celeb1}-Fig.~\ref{figure:churches1}.



\section{Intuition behind Our Ideas}
\subsection{Intermediate Results of Reverse Diffusion Process}
We provide the intermediate results of generation with the DDIM scheduler in Fig.~\ref{figure:ddim_process}. We visualize the predicted $x_0$ at each timestep, which is mentioned in Eq. 12 of~\cite{song2020denoising}. From this visualization results, we can say that the diffusion models can lower the domain discrepancy and be helpful to the matching process.

\section{Limitations}
One obvious limitation that MIDMs possess is slow sampling speed due to the characteristic of the diffusion model. While diffusion models generate plausible samples and DDIM sampler~\cite{song2020denoising} improves the sampling speed, sampling is still slower than other generative models like GANs~\cite{goodfellow2014generative}. One straightforward solution would be to consider combining ours with recent sampling acceleration approaches~\cite{liu2022pseudo,watson2021learning,nichol2021improved}.
Our model causes the increment of the computational cost because of the added diffusion model and the encoder-decoder. This makes the training more difficult because it makes the necessity of the larger memory and faster GPU. We would address this problem by optimizing the model size by some search of model hyper-parameter or using mixed-precision training~\cite{micikevicius2017mixed}.

\section{Broader Impact}
MIDMs enable the generation of high-quality images while faithfully bringing the local style of the exemplars, and utilizing these strengths, MIDMs can be used for a variety of applications, such as image editing and style transfer. On the other hand, our model risks being used for malicious works, such as deep fakes. Also, since the model learns to approximate the distribution of a training dataset, it can model the same bias that the training sets have, such as gender, race, and age.

\include{Figure/celeb_1}
\include{Figure/fashion_1}
\include{Figure/churches_1}
\input{Figure/intermediate_ddim}

%% file: Table/table_hyperparameters.tex
\begin{table*}[h!]

\small
\centering
\caption{\textbf{List of MIDMs hyperparameters for CelebA-HQ~\cite{liu2015deep} and DeepFashion~\cite{liu2016deepfashion}}}\label{table:hyperparameter} \vspace{-5pt}
\begin{tabular}{cccc}
\toprule
{hyperparameter} & CelebA-HQ~\cite{liu2015deep} & DeepFashion~\cite{liu2016deepfashion}
\\
\midrule
$S$ & 16 & 16 \\
$T$ & 4  & 4 \\
$\gamma$ & 0.3 & 0.5 \\
$\lambda_{\mathrm{perc}}$ & 0.002 & 0.002 \\
$\lambda_{\mathrm{src}}$ & 1.0 & 1.0 \\
$\lambda_{\mathrm{style}}$ & 1.0 & 1.0 \\
$\lambda_{\mathrm{cycle}}$ & 1.0 & 1.0 \\
$\lambda_{\mathrm{dom}}$ & 10.0 & 10.0 \\
$\lambda_{\mathrm{diff}}$ & 1.0 & 1.0 \\
\bottomrule
\end{tabular}
\hspace{1em}
\end{table*} 

%% file: Table/sampling_algorithm.tex
\begin{algorithm}[h]
\caption{MIDMs' sampling, given a diffusion model $\epsilon_{\theta}(x_t)$, pretrained encoder $\mathcal{E}$, pretrained decoder $\mathcal{D}$, feature extractor  $\mathcal{F}_\mathcal{X}$ for condition,  $\mathcal{F}_\mathcal{Y}$ for exemplar and   $\mathcal{F}^\mathrm{iter}_\mathcal{Y}$ for recurrent matching.
}\label{alg:sample_alg}
\begin{algorithmic}[1]
    \State \textbf{Input}: conditional image $I_\mathcal{X}$ and exemplar image $I_\mathcal{Y}$
    \State \textbf{Output}: generated image $I_{\mathcal{X}\leftarrow \mathcal{Y}}$
    \State $\mathrm{warp}(\cdot)$ : soft-warping function stated in Eq.6
    \State $\mathrm{confidence\_mask}(\cdot)$ : confidence mask stated in Eq.10
    \State $D_\mathcal{X},D_\mathcal{Y} \gets \mathcal{E}(I_\mathcal{X}), \mathcal{E}(I_\mathcal{Y})$
    \State $S_\mathcal{X}, S_\mathcal{Y} \gets \mathcal{F}_\mathcal{X}(D_\mathcal{X}), \mathcal{F}_\mathcal{Y}(D_\mathcal{Y})$
    \State $C_{\mathcal{X}\leftarrow \mathcal{Y}} \gets \mathrm{normalize}(S_\mathcal{X}) \cdot \mathrm{normalize}(S_\mathcal{Y})^{T}$
    \State $R_{\mathcal{X}\leftarrow \mathcal{Y}} \gets \mathrm{warp}(C_{\mathcal{X}\leftarrow \mathcal{Y}},D_\mathcal{Y})$
    \State $\epsilon \gets \text{sample from } \mathcal{N}(0, \mathbf{I})$
    \State $r^{N}_{\mathcal{Y}} \gets \sqrt{\alpha_{\tau_N}} R_{\mathcal{X}\leftarrow \mathcal{Y}} + \sqrt{1 - \alpha_{\tau_N}} \epsilon$
    \ForAll{$n$ from $N$ to 2}
        \State $\Tilde{r}^{n}_{\mathcal{Y}} \gets f_{\theta}(r^{n}_{\mathcal{Y}},\tau_n)$
        \State $ S^\mathrm{iter}_\mathcal{Y} \gets \mathcal{F}^\mathrm{iter}_\mathcal{Y}(\Tilde{r}^{n}_{\mathcal{Y}},D_\mathcal{X})$
        \State $C^\mathrm{iter}_{\mathcal{X}\leftarrow \mathcal{Y}} \gets  \mathrm{normalize}(S^\mathrm{iter}_\mathcal{Y}) \cdot \mathrm{normalize}(S_\mathcal{Y})^{T}$
        \State $R_{\Tilde{r}^{n}_{\mathcal{Y}}\leftarrow \mathcal{Y}} \gets \mathrm{warp}(C^\mathrm{iter}_{\mathcal{X}\leftarrow \mathcal{Y}},D_\mathcal{Y})$
        \State $M_{\Tilde{r}^{n}_{\mathcal{Y}}\leftarrow \mathcal{Y}} \gets \mathrm{confidence\_mask}(C^\mathrm{iter}_{\mathcal{X}\leftarrow \mathcal{Y}})$ 
        \State $r^{n-1}_{\mathcal{Y}} \gets \sqrt{\alpha_{\tau_{n-1}}} 
(M_{\Tilde{r}^{n}_{\mathcal{Y}}\leftarrow \mathcal{Y}} \odot R_{\Tilde{r}^{n}_{\mathcal{Y}}\leftarrow \mathcal{Y}} + (1-M_{\Tilde{r}^{n}_{\mathcal{Y}}\leftarrow \mathcal{Y}}) \odot \Tilde{r}^{n}_{\mathcal{Y}}) + \sqrt{1 - \alpha_{\tau_{n-1}}} \epsilon_{\theta}(r^{n}_{\mathcal{Y}},\tau_{n})$
    \EndFor
    \State $r^{0}_{\mathcal{Y}} \gets f_{\theta}(r^{1}_{\mathcal{Y}})$
    \State $I_{\mathcal{X}\leftarrow \mathcal{Y}} \gets \mathcal{D}(r^{0}_{\mathcal{Y}})$ \\
    \Return $I_{\mathcal{X}\leftarrow \mathcal{Y}}$
\end{algorithmic}
\end{algorithm}

%% file: Table/sampling_step_table.tex
\begin{table*}[h!]

\small
\centering
\caption{\textbf{Ablation study on the number of sampling steps.} We validate the performance on CelebA-HQ~\cite{liu2015deep}.}\label{table:sampling_steps} \vspace{-5pt}
\begin{tabular}{lccc}
\toprule
{Models} & {FID}$\downarrow$ & {SWD}$\downarrow$ 
\\
\midrule
$T=200, N=50$     & \textbf{15.67} & \textbf{12.34}    \\
$T=100, N=25$      & 16.99 & 13.33   \\
$T=16, N=4$        & 17.94 & 13.99 \\
\bottomrule
\end{tabular}
\hspace{1em}
\end{table*} 

%% file: Figure/gan_abl.tex
\begin{table*}
\centering
\setlength\tabcolsep{0pt}
\begin{subtable}[h]{1.0\textwidth}
    \renewcommand{\arraystretch}{0.0}
    \begin{tabular}{@{}ccccc@{}}

        \includegraphics[width=0.2\columnwidth]{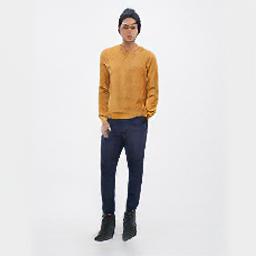}&
        \includegraphics[width=0.2\columnwidth]{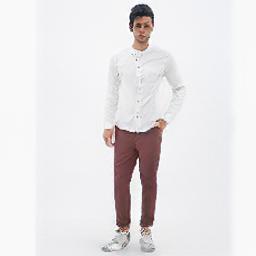}&
        \includegraphics[width=0.2\columnwidth]{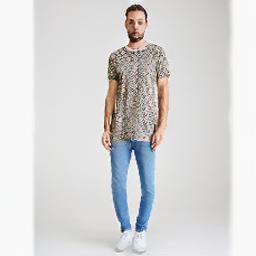}&
        \includegraphics[width=0.2\columnwidth]{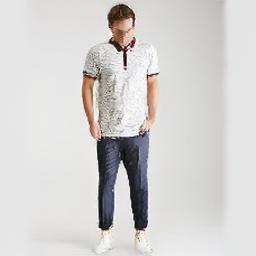}&
        \includegraphics[width=0.2\columnwidth]{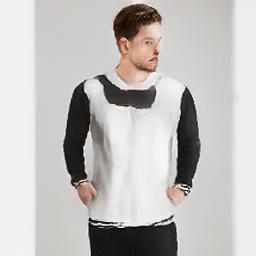}\\

        \includegraphics[width=0.2\columnwidth]{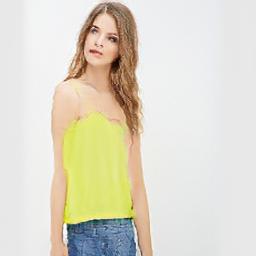}&
        \includegraphics[width=0.2\columnwidth]{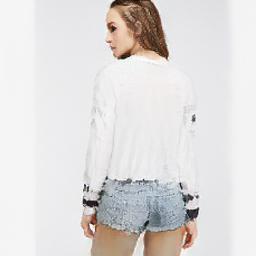}&
        \includegraphics[width=0.2\columnwidth]{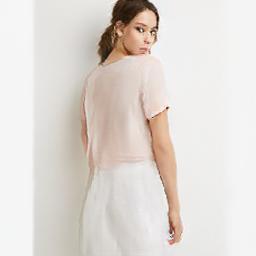}&
        \includegraphics[width=0.2\columnwidth]{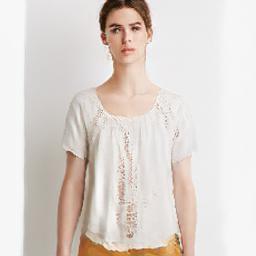}&
        \includegraphics[width=0.2\columnwidth]{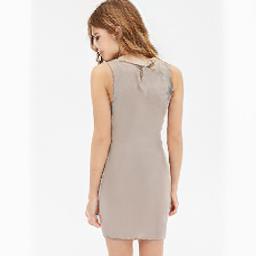}\\ \\
    \end{tabular}
    \caption{\textbf{Generated by original CoCosNet~\cite{zhang2020cross} generator.}}
\end{subtable}
\begin{subtable}[h]{1.0\textwidth}
    \renewcommand{\arraystretch}{0.0}
    \begin{tabular}{@{}ccccc@{}}

        \includegraphics[width=0.2\columnwidth]{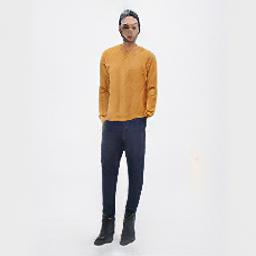}&
        \includegraphics[width=0.2\columnwidth]{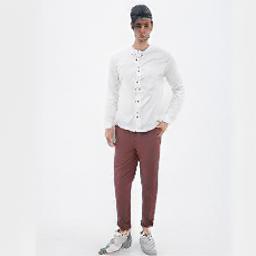}&
        \includegraphics[width=0.2\columnwidth]{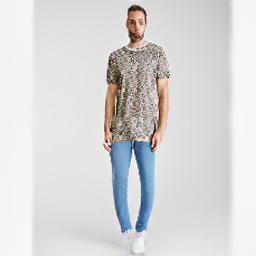}&
        \includegraphics[width=0.2\columnwidth]{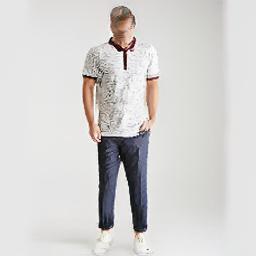}&
        \includegraphics[width=0.2\columnwidth]{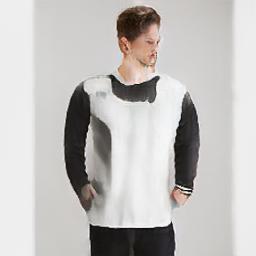}\\
    
        \includegraphics[width=0.2\columnwidth]{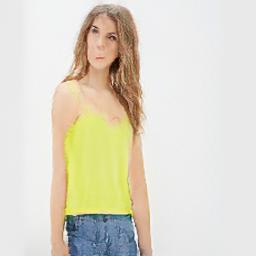}&
        \includegraphics[width=0.2\columnwidth]{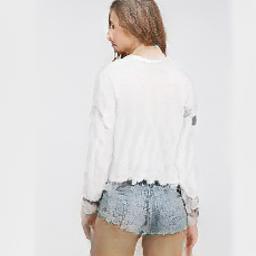}&
        \includegraphics[width=0.2\columnwidth]{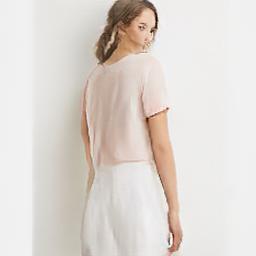}&
        \includegraphics[width=0.2\columnwidth]{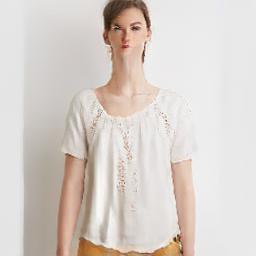}&
        \includegraphics[width=0.2\columnwidth]{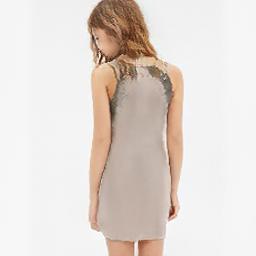}\\
    
    \end{tabular}
    \caption{\textbf{Iterative matching-and-generation results by CoCosNet~\cite{zhang2020cross} generator.}}
\end{subtable}
\caption{\textbf{Comparisions between generation results of original CoCosNet and CoCosNet with iterative matching-and-generation.}}
\label{figure:gan_abl}
\end{table*}


%% file: Table/gan_ablation.tex
\renewcommand\arraystretch{1.0}
\begin{table}[h!]
\caption{
\textbf{Ablation study of GAN-based iterative matching-and-generation framework on DeepFashion~\cite{liu2016deepfashion} datasets.}\\ 
}\label{tab:gan_abl}
\vspace{-5pt}
\small
\centering 
\begin{tabular}{l|cc} \toprule
Methods & FID \\ \midrule
CoCosNet~\cite{zhang2020cross}                                              & \textbf{14.4}  \\
CoCosNet w/iterative matching \& generation~\cite{zhang2020cross}         & 19.05  \\ 
\bottomrule
\end{tabular} 
\end{table}

%% file: Figure/celeb_1.tex
\begin{figure*}[t]
\center
\small
\setlength\tabcolsep{0pt}
{
\renewcommand{\arraystretch}{0.0}
\begin{tabular}{@{}rrccccc@{}}
    &
    \raisebox{0.5\height}{\rotatebox{90}{Exemplars}}&
    \includegraphics[width=0.2\columnwidth]{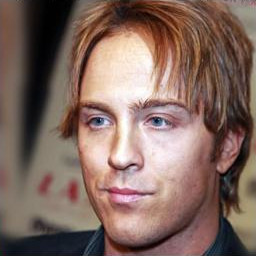}&
    \includegraphics[width=0.2\columnwidth]{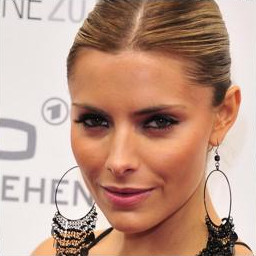}&
    \includegraphics[width=0.2\columnwidth]{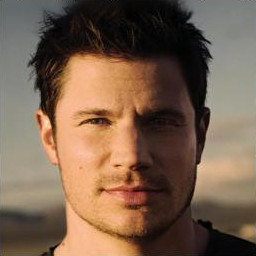}&
    \includegraphics[width=0.2\columnwidth]{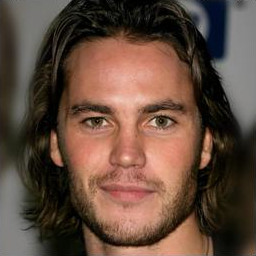}\\

    \raisebox{0.5\height}{\rotatebox{90}{Conditions}}&
    \includegraphics[width=0.2\columnwidth]{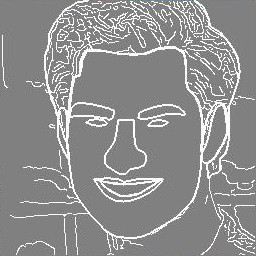}&
    \includegraphics[width=0.2\columnwidth]{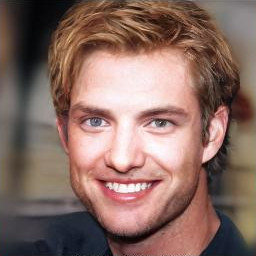}&
    \includegraphics[width=0.2\columnwidth]{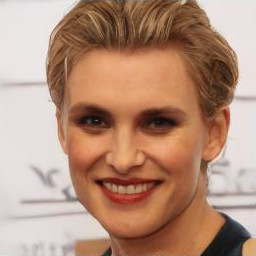}&
    \includegraphics[width=0.2\columnwidth]{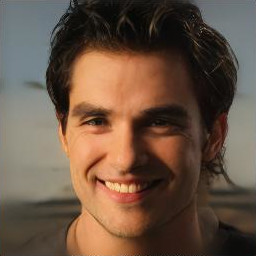}&
    \includegraphics[width=0.2\columnwidth]{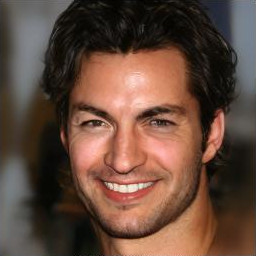}\\
\end{tabular}
\begin{tabular}{@{}rrccccc@{}}
    &
    \raisebox{0.5\height}{\rotatebox{90}{Exemplars}}&
    \includegraphics[width=0.2\columnwidth]{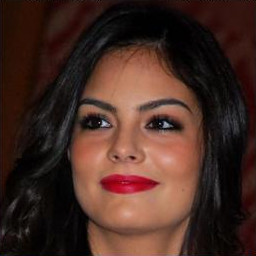}&
    \includegraphics[width=0.2\columnwidth]{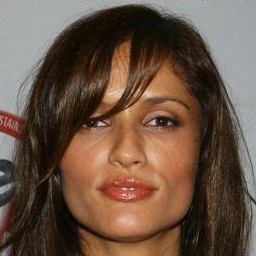}&
    \includegraphics[width=0.2\columnwidth]{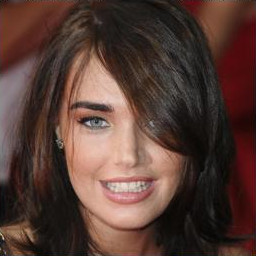}&
    \includegraphics[width=0.2\columnwidth]{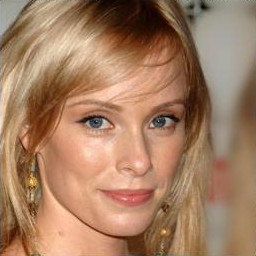}\\

    \raisebox{0.5\height}{\rotatebox{90}{Conditions}}&
    \includegraphics[width=0.2\columnwidth]{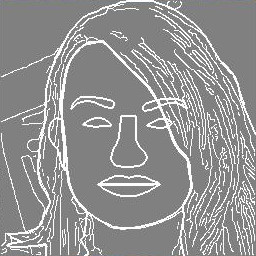}&
    \includegraphics[width=0.2\columnwidth]{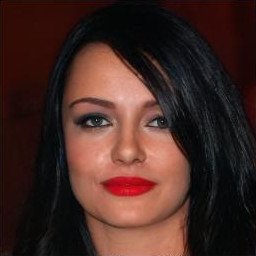}&
    \includegraphics[width=0.2\columnwidth]{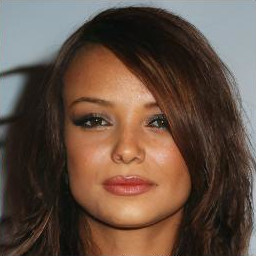}&
    \includegraphics[width=0.2\columnwidth]{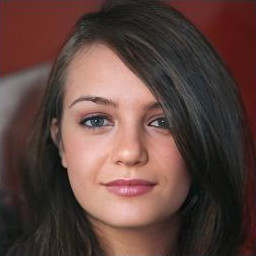}&
    \includegraphics[width=0.2\columnwidth]{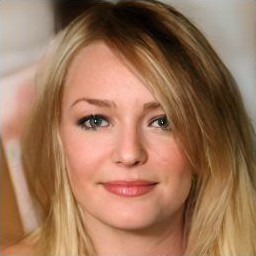}\\
\end{tabular}
\begin{tabular}{@{}rrccccc@{}}
    &
    \raisebox{0.5\height}{\rotatebox{90}{Exemplars}}&
    \includegraphics[width=0.2\columnwidth]{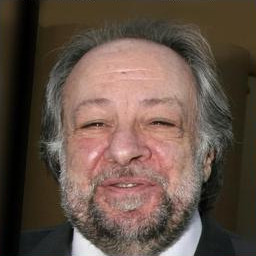}&
    \includegraphics[width=0.2\columnwidth]{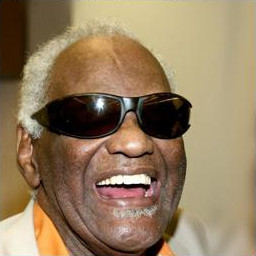}&
    \includegraphics[width=0.2\columnwidth]{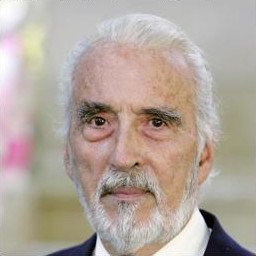}&
    \includegraphics[width=0.2\columnwidth]{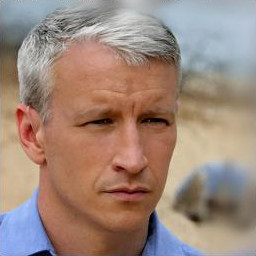}\\

    \raisebox{0.5\height}{\rotatebox{90}{Conditions}}&
    \includegraphics[width=0.2\columnwidth]{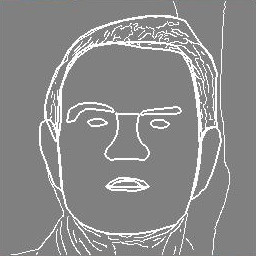}&
    \includegraphics[width=0.2\columnwidth]{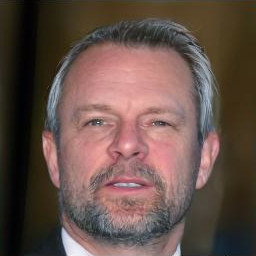}&
    \includegraphics[width=0.2\columnwidth]{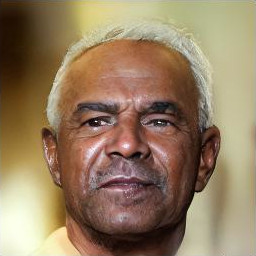}&
    \includegraphics[width=0.2\columnwidth]{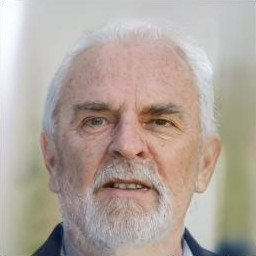}&
    \includegraphics[width=0.2\columnwidth]{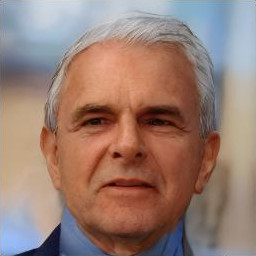}\\
\end{tabular}
}
\caption{\textbf{Qualitative results on CelebA-HQ~\cite{liu2015deep}}}
\label{figure:celeb1}
\end{figure*}

%% file: Figure/fashion_1.tex
\begin{figure*}[t]
\center
\small
\setlength\tabcolsep{0pt}
{
\renewcommand{\arraystretch}{0.0}
\begin{tabular}{@{}rrccccc@{}}
    &
    \raisebox{0.5\height}{\rotatebox{90}{Exemplars}}&
    \includegraphics[width=0.2\columnwidth]{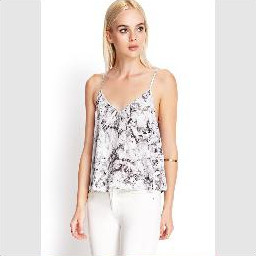}&
    \includegraphics[width=0.2\columnwidth]{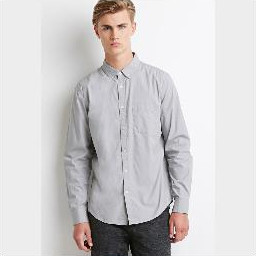}&
    \includegraphics[width=0.2\columnwidth]{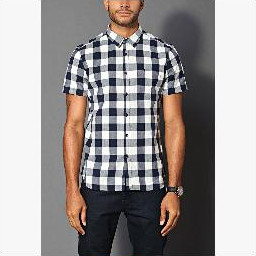}&
    \includegraphics[width=0.2\columnwidth]{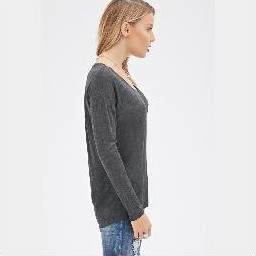}\\

    \raisebox{0.5\height}{\rotatebox{90}{Conditions}}&
    \includegraphics[width=0.2\columnwidth]{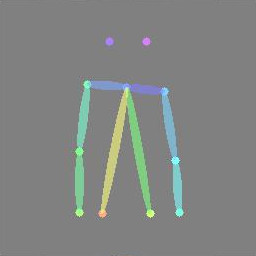}&
    \includegraphics[width=0.2\columnwidth]{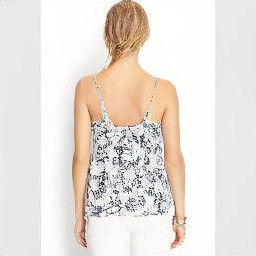}&
    \includegraphics[width=0.2\columnwidth]{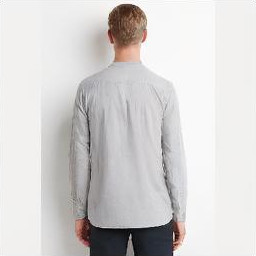}&
    \includegraphics[width=0.2\columnwidth]{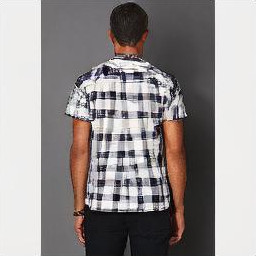}&
    \includegraphics[width=0.2\columnwidth]{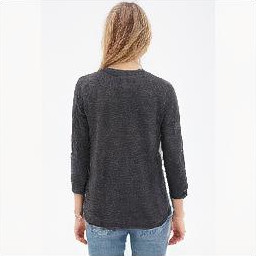}\\
\end{tabular}
\begin{tabular}{@{}rrccccc@{}}
    &
    \raisebox{0.5\height}{\rotatebox{90}{Exemplars}}&
    \includegraphics[width=0.2\columnwidth]{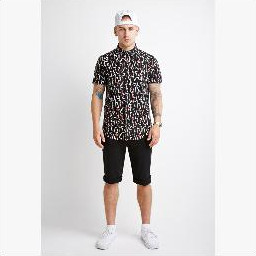}&
    \includegraphics[width=0.2\columnwidth]{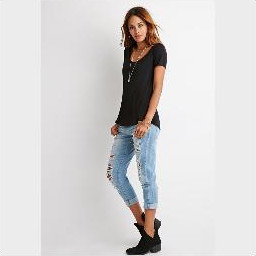}&
    \includegraphics[width=0.2\columnwidth]{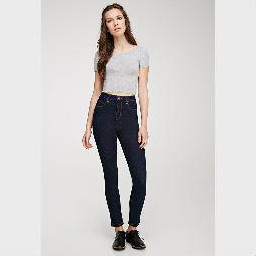}&
    \includegraphics[width=0.2\columnwidth]{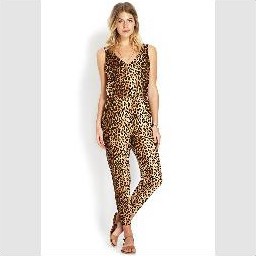}\\

    \raisebox{0.5\height}{\rotatebox{90}{Conditions}}&
    \includegraphics[width=0.2\columnwidth]{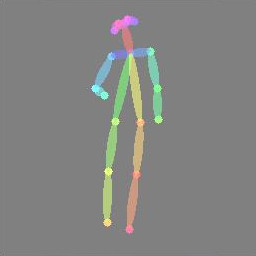}&
    \includegraphics[width=0.2\columnwidth]{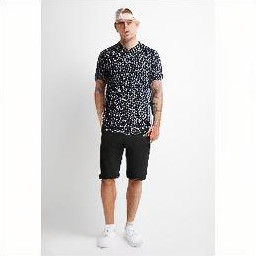}&
    \includegraphics[width=0.2\columnwidth]{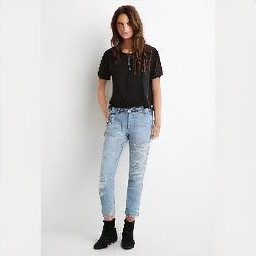}&
    \includegraphics[width=0.2\columnwidth]{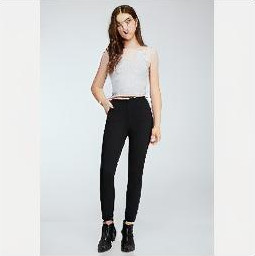}&
    \includegraphics[width=0.2\columnwidth]{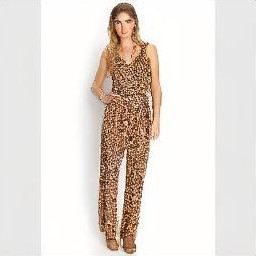}\\
\end{tabular}
\begin{tabular}{@{}rrccccc@{}}
    &
    \raisebox{0.5\height}{\rotatebox{90}{Exemplars}}&
    \includegraphics[width=0.2\columnwidth]{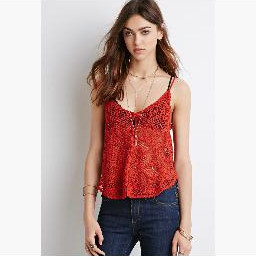}&
    \includegraphics[width=0.2\columnwidth]{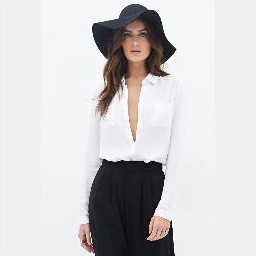}&
    \includegraphics[width=0.2\columnwidth]{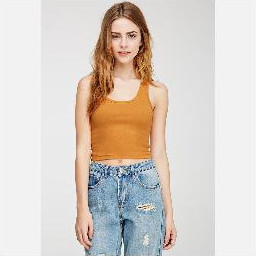}&
    \includegraphics[width=0.2\columnwidth]{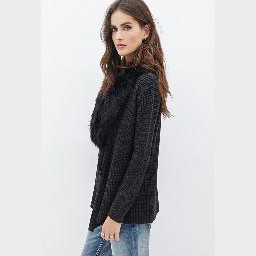}\\

    \raisebox{0.5\height}{\rotatebox{90}{Conditions}}&
    \includegraphics[width=0.2\columnwidth]{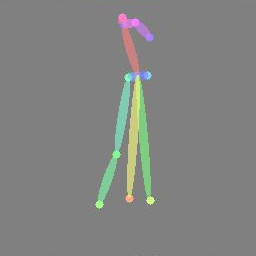}&
    \includegraphics[width=0.2\columnwidth]{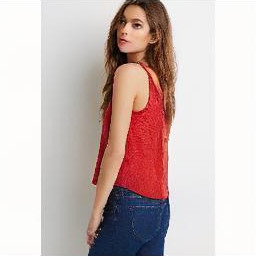}&
    \includegraphics[width=0.2\columnwidth]{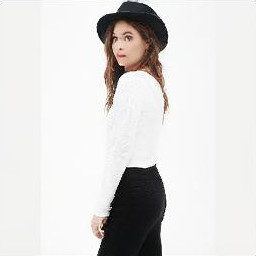}&
    \includegraphics[width=0.2\columnwidth]{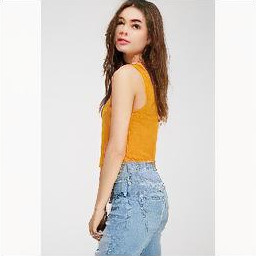}&
    \includegraphics[width=0.2\columnwidth]{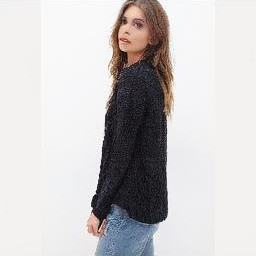}\\
\end{tabular}
}
\caption{\textbf{Qualitative results on DeepFashion~\cite{liu2016deepfashion}}}
\label{figure:fashion1}
\end{figure*}

%% file: Figure/churches_1.tex
\begin{figure*}[t]
\center
\small
\setlength\tabcolsep{0pt}
{
\renewcommand{\arraystretch}{0.0}

\begin{tabular}{@{}rrccccc@{}}
    &
    \raisebox{0.5\height}{\rotatebox{90}{Exemplars}}&
    \includegraphics[width=0.2\columnwidth]{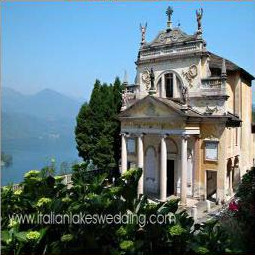}&
    \includegraphics[width=0.2\columnwidth]{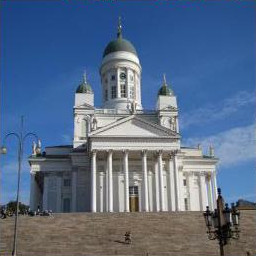}&
    \includegraphics[width=0.2\columnwidth]{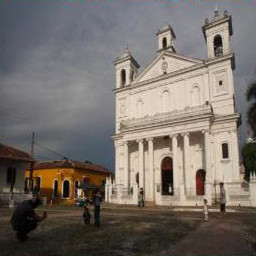}&
    \includegraphics[width=0.2\columnwidth]{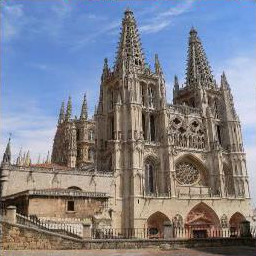}\\

    \raisebox{0.5\height}{\rotatebox{90}{Conditions}}&
    \includegraphics[width=0.2\columnwidth]{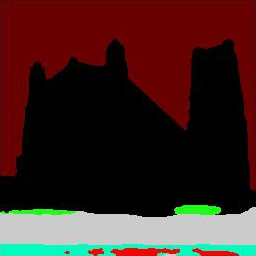}&
    \includegraphics[width=0.2\columnwidth]{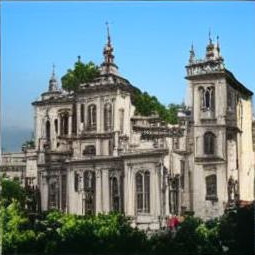}&
    \includegraphics[width=0.2\columnwidth]{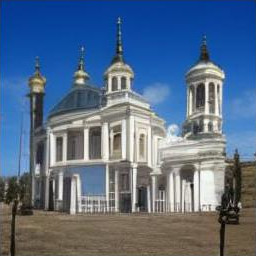}&
    \includegraphics[width=0.2\columnwidth]{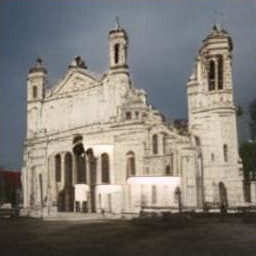}&
    \includegraphics[width=0.2\columnwidth]{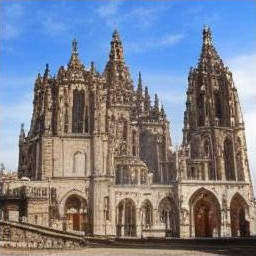}\\
\end{tabular}

\begin{tabular}{@{}rrccccc@{}}
    &
    \raisebox{0.5\height}{\rotatebox{90}{Exemplars}}&
    \includegraphics[width=0.2\columnwidth]{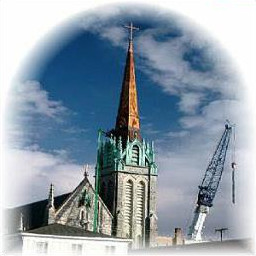}&
    \includegraphics[width=0.2\columnwidth]{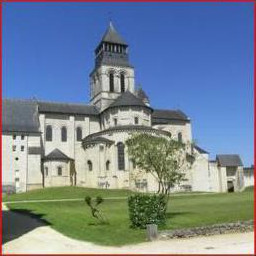}&
    \includegraphics[width=0.2\columnwidth]{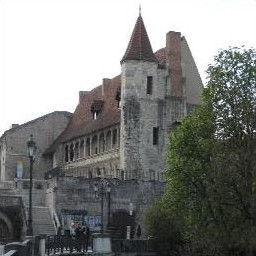}&
    \includegraphics[width=0.2\columnwidth]{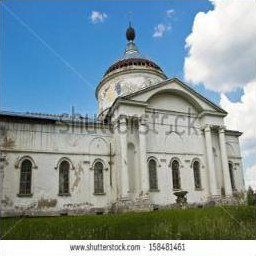}\\

    \raisebox{0.5\height}{\rotatebox{90}{Conditions}}&
    \includegraphics[width=0.2\columnwidth]{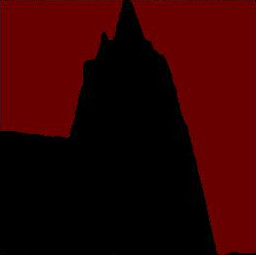}&
    \includegraphics[width=0.2\columnwidth]{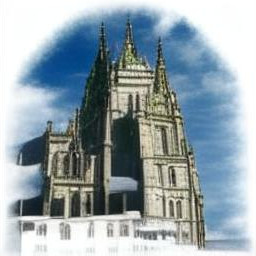}&
    \includegraphics[width=0.2\columnwidth]{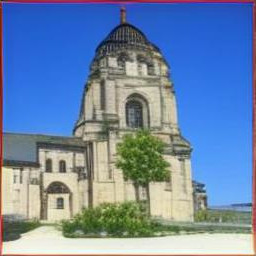}&
    \includegraphics[width=0.2\columnwidth]{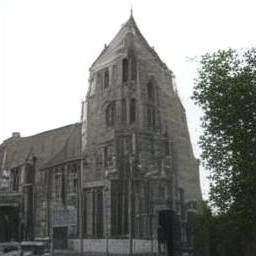}&
    \includegraphics[width=0.2\columnwidth]{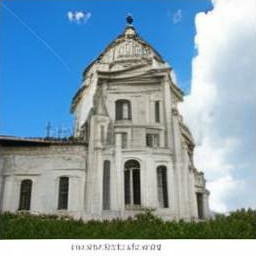}\\
\end{tabular}
}
\caption{\textbf{Qualitative results on LSUN-Churches~\cite{yu2015lsun}} We use segmentation map generated by ADE20k~\cite{zhou2017scene}-pretrained Swin-S~\cite{liu2021swin} model.}
\label{figure:churches1}
\end{figure*}

%% file: Figure/intermediate_ddim.tex
\begin{figure*}[h]
\center
\small
\setlength\tabcolsep{0pt}
{
\renewcommand{\arraystretch}{1.0}
\begin{tabular}{@{}ccccc@{}}
    \includegraphics[width=0.2\columnwidth]{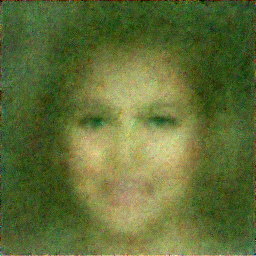}&
    \includegraphics[width=0.2\columnwidth]{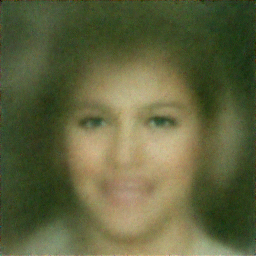}&
    \includegraphics[width=0.2\columnwidth]{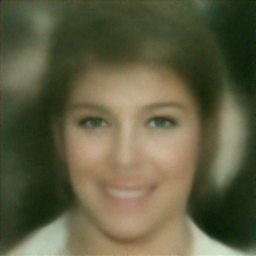}&
    \includegraphics[width=0.2\columnwidth]{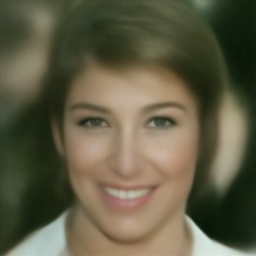}&
    \includegraphics[width=0.2\columnwidth]{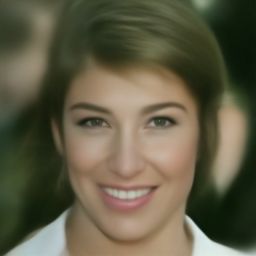}\\
    $t=25$ & $t=50$ & $t=75$ & $t=100$ & $t=125$\\

    \includegraphics[width=0.2\columnwidth]{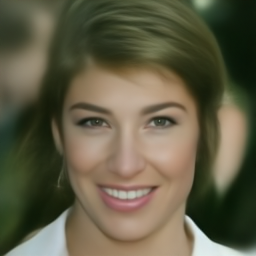}&
    \includegraphics[width=0.2\columnwidth]{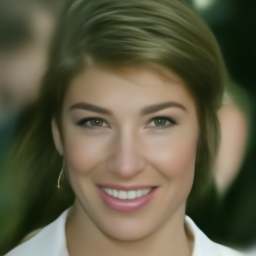}&
    \includegraphics[width=0.2\columnwidth]{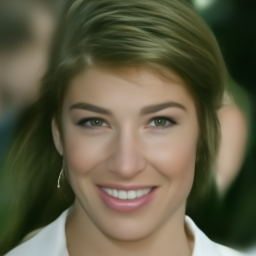}&
    \includegraphics[width=0.2\columnwidth]{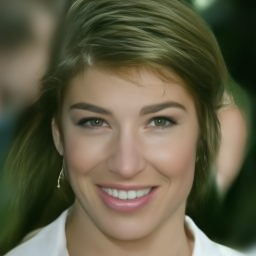}&
    \includegraphics[width=0.2\columnwidth]{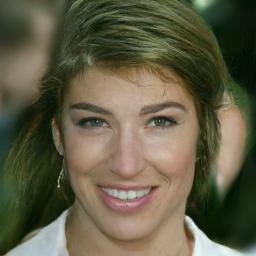}\\
    $t=150$ & $t=175$ & $t=200$ & $t=225$ & $t=250$\\ \\
\end{tabular}
}
{
\renewcommand{\arraystretch}{1.0}
\begin{tabular}{@{}ccccc@{}}
    \includegraphics[width=0.2\columnwidth]{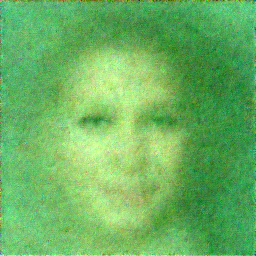}&
    \includegraphics[width=0.2\columnwidth]{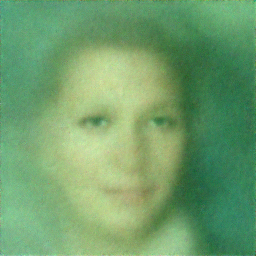}&
    \includegraphics[width=0.2\columnwidth]{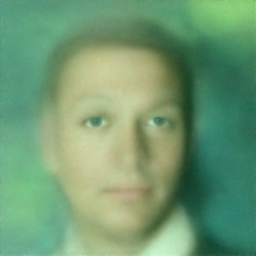}&
    \includegraphics[width=0.2\columnwidth]{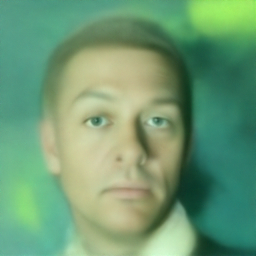}&
    \includegraphics[width=0.2\columnwidth]{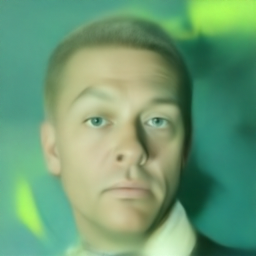}\\
    $t=25$ & $t=50$ & $t=75$ & $t=100$ & $t=125$\\

    \includegraphics[width=0.2\columnwidth]{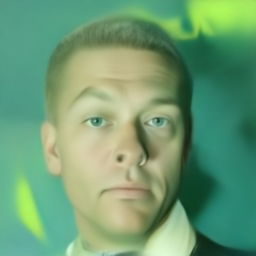}&
    \includegraphics[width=0.2\columnwidth]{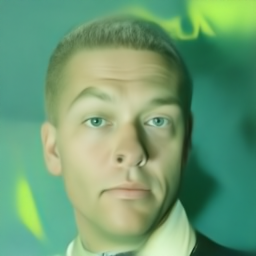}&
    \includegraphics[width=0.2\columnwidth]{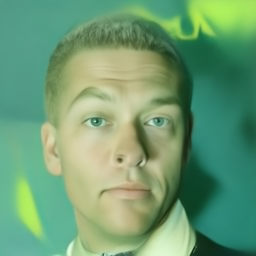}&
    \includegraphics[width=0.2\columnwidth]{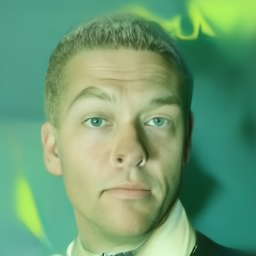}&
    \includegraphics[width=0.2\columnwidth]{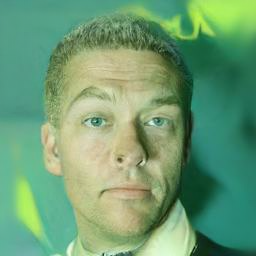}\\
    $t=150$ & $t=175$ & $t=200$ & $t=225$ & $t=250$\\
\end{tabular}
}
\caption{\textbf{Progressive results of predicted $x_0$ directly from $x_t$ at each timestep $t$ when sampling from the DDIM scheduler~\cite{song2020denoising}.}}
\label{figure:ddim_process}
\end{figure*}